\UseRawInputEncoding
\documentclass[journal]{IEEEtran}
\IEEEoverridecommandlockouts
\usepackage{cite}
\usepackage{verbatim}
\usepackage{float}
\usepackage{dblfloatfix}
\usepackage{amsmath,amssymb,amsfonts}
\usepackage{algorithmic}
\usepackage{graphicx}
\usepackage{textcomp}
\usepackage{xcolor}
\colorlet{red}{red}
\colorlet{magenta}{magenta}
\usepackage{makecell}
\usepackage{enumitem,kantlipsum}
\usepackage{colortbl}
\usepackage{graphicx}
\usepackage{adjustbox}
\usepackage{hyperref}
\usepackage{longtable}
\usepackage{booktabs}
\usepackage{multirow}
\def\BibTeX{{\rm B\kern-.05em{\sc i\kern-.025em b}\kern-.08em
    T\kern-.1667em\lower.7ex\hbox{E}\kern-.125emX}}

\begin{document}

\title{Energy- and Memory-Efficient PEFT Methods for Personalized On-Device SLMs on Consumer GPUs} 

\author{Kuanysh~Akhmetzhanov and Jurn-Gyu Park
\thanks{Corresponding author: Jurn-Gyu Park (jurn.park@nu.edu.kz).}%
\thanks{K. Akhmetzhanov and J-.G. Park are with the Department of Computer Science, Nazarbayev University, Astana 010000, Kazakhstan (e-mail: kuanysh.akhmetzhanov@nu.edu.kz, jurn.park@nu.edu.kz).}
\thanks{The authors gratefully acknowledge Saltanat Ami for her editorial assistance and valuable feedback on this manuscript.}%

}

\markboth{Preprint, 2026}%
{Akhmetzhanov: Energy- and Memory-Efficient PEFT Methods for Personalized On-Device SLMs on Consumer GPUs} 

\maketitle


{\color{black}

\begin{abstract}
\noindent
Despite rapid advances in large language models (LLMs), deploying and personalizing them on resource-constrained devices remains impractical due to high VRAM, time, and energy costs. Parameter-Efficient Fine-Tuning (PEFT) of Small Language Models (SLMs) offers a promising alternative, yet few studies compare PEFT methods across architectures using both general and personalization benchmarks while accounting for energy consumption. We compare five fine-tuning approaches (Full Fine-Tuning, LoRA, LoRA+, QLoRA, and BitFit) on four SLMs from two families (Transformer-based: TinyLlama-1.1B, Qwen3-1.7B; SSM-based: Mamba-1.4B, Mamba-2-1.3B) across three GLUE tasks (SST-2, QNLI, STS-B) and three LaMP personalization tasks (LaMP-1, LaMP-2, LaMP-3). Each configuration is evaluated with the energy-focused NetScore-E and the memory-focused NetScore-M, the two variants that reflect the constraints binding on-device deployment. Methods are selected with a strict energy-first rule (highest NetScore-E, ties broken by NetScore\#). LoRA+ achieves the highest NetScore-E in 19 of 24 configurations and the highest NetScore-M in 13 of 24, and is the selected method in 18 of 24. QLoRA, available only for the Transformer models, cuts peak finetuning VRAM by up to $3.9\times$ relative to LoRA and therefore takes the best NetScore-M in 5 of the 12 Transformer configurations, although its de-quantization overhead leaves it selected in only one of them once energy decides. BitFit and full fine-tuning are almost never competitive on either variant, and TinyLlama-1.1B leads the energy-focused NetScore-E on five of the six benchmarks and the memory-focused NetScore-M on four. These results show that compact SLMs paired with PEFT provide a practical, energy-aware path to personalized on-device deployment, with the optimal method set by the dominant constraint: LoRA+ for energy and QLoRA for memory.

\end{abstract}
}


\IEEEpeerreviewmaketitle


\section{Introduction}


{\color{black}
\IEEEPARstart{T}{here} have been
major developments in the area of Natural Language Processing (NLP) during the last ten years due to the increasing scalability of neural language models. It started with the emergence of the Transformer architecture~\cite{vaswani2023attentionneed}, and then with the large-scale pre-training models of BERT~\cite{devlin2019bertpretrainingdeepbidirectional} and GPT~\cite{yenduri2023generativepretrainedtransformercomprehensive}. With that came the "larger is better" trend~\cite{hoffmann2022trainingcomputeoptimallargelanguage},
which led to the exponential increase in model sizes from hundreds of millions of parameters to hundreds of billions.
This resulted in LLMs that could generate code, reason for multiple steps and engage users in open ended dialogues with extremely high levels of accuracy. 
While there was an enormous amount of attention given to LLMs that could perform complex reasoning, there also existed a growing demand towards on-device and personalized AI where language models will operate independently on a consumer single GPU, smart phones, embedded platforms, and various types of edge computing hardware. They would need to adapt to the unique needs and preferences of individual users without having access to remote cloud resources.

These two trends represent contrasting ideas and philosophies. While state of the art LLMs can often require hundreds of GB memory to operate efficiently\cite{pan2025costbenefitanalysisonpremiselarge}, they cannot effectively be deployed on consumer grade GPUs, let alone on smaller mobile or embedded systems. To support such deployments and fine-tuning, organizations may be required to utilize expensive cloud APIs or proprietary servers to provide the necessary computational capacity to run large language models. Each option creates additional costs for latency, privacy and operational expenses that create barriers for widespread adoption\cite{irugalbandara2024scalingscaleupcostbenefit}. 
Additionally, the fine-tuning process for large language models further complicates
many of these issues. Updating billions of parameters
requires increased amounts of VRAM, longer periods of training time and greater amounts of energy usage\cite{xia2024understandingperformanceestimatingcost}. All of these factors combine to make large language models difficult to deploy and personalize with fine-tuning in resource-constrained environments.

To help bridge this gap between the abilities of current frontier LLMs and those that can currently be realistically used and personalized within resource constraints, researchers have proposed two approaches. 
One method involves developing what are referred to as SLMs, compact architectures with model sizes ranging from 1 -- 2 Billion parameters,
which when trained on very large datasets containing many diverse samples, can produce similar results to larger models on specific tasks\cite{belcak2025smalllanguagemodelsfuture,nguyen-etal-2025-survey,schick-schutze-2021-just,lu2025smalllanguagemodelssurvey,subramanian2025smalllanguagemodelsslms}. Examples of SLMs include: TinyLlama-1.1B and Qwen3-1.7B, which are members of the transformer family, and Mamba-1.4B and Mamba-2-1.3B, which are structured state space models (SSMs), which offer linear time sequence processing as an alternative to the quadratic time complexity associated with self-attention. 

Another strategy involves Parameter Efficient Fine Tuning (PEFT). This strategy modifies only a fraction of the parameters in a pre-trained model. Examples include: LoRA, which adds low rank updates to the weights, LoRA+, an enhanced version of LoRA, QLoRA, a quantized version of LoRA, and BitFit, which only updates the biases of the model. Many recent studies have demonstrated how PEFT can achieve competitive results while significantly reducing memory and compute requirements for fine-tuning large language models\cite{pu2023empiricalanalysisstrengthsweaknesses,clarke2024peftuparameterefficientfinetuninguser,balne2024parameterefficientfinetuning,zhang2025parameterefficientfinetuningfoundationmodels}. 

}{\color{black}
A key limitation however is that most studies that investigate PEFT parameters (not VRAM sizes) do so based on one type of architecture family (e.g., transformers), one or two benchmark sets  or do not account for trade offs related to energy efficiency alongside improvements in task accuracy. Secondly, evaluations typically rely on  general language understanding benchmark tasks, often neglecting personalization tasks, despite the fact that adapting a model to individual user preferences is a primary motivation for on-device fine-tuning~\cite{clarke2024peftuparameterefficientfinetuninguser}. Additionally, most studies report only task accuracy and memory footprint, while failing to report the energy consumed during training, which is one of the most important constraints for battery powered deployments.

}{\color{black}
To do this, we develop a comprehensive experimental framework that unifies both strategies (SLMs + PEFTs) and examines them together across metrics including performance, efficiency, and sustainability. 
We employed full fine-tuning as well as four PEFT methods (LoRA, LoRA+, QLoRA and BitFit) to fine-tune four SLMs from two architectural families (Transformer-based: TinyLlama-1.1B and Qwen3-1.7B, SSM-based: Mamba-1.4B and Mamba-2-1.3B). 
Each configuration was evaluated on three General Language Understanding Evaluation (GLUE) benchmarks (SST-2, QNLI, STS-B) for general language understanding, and three LaMP benchmarks (LaMP-1, LaMP-2, LaMP-3) for personalization. 
This gives 24 model--task configurations and 108 fine-tuning runs in total, as QLoRA is not supported for the two SSM models. Alongside traditional evaluation metrics (i.e., test accuracy), we measured training time, VRAM usage, average power consumption, total energy consumption, and TFLOPs. Those measurements then were used to calculate a metric called NetScore\cite{wong2018netscoreuniversalmetricslargescale,wong2019attonetscompactefficientdeep}, alongside its extensions: NetScore-E\cite{trinci2024greencontinuallearningreally,wang2026attentionbasedfeaturememorydesign}, NetScore-M~\cite{toktassyn2026efficient}, NetScore\#~\cite{toktassyn2026efficient}. \\
%
}

{\color{black}
The main contributions of this paper are as follows:
\begin{itemize}
\item Demonstrate various PEFT methods on SLMs to match or surpass full fine-tuning performance on multiple tasks. 

\item Provide empirical comparisons of Transformer- and SSM-based models, analyzing their trade-offs across accuracy, energy consumption, memory usage, and training time on both general GLUE and personalized LaMP benchmarks.

\item Utilize multiple quantitative metrics and NetScores, composite score metrics that balance task performance against efficiency.

\item Publicly release all training and evaluation scripts for reproducibility of the results.\footnote{Codes at: \url{https://github.com/Rickserd/super-duper-computing-machine/}} 
\end{itemize}
}

\section{Motivation and Related Work}

\subsection{Motivation}


\subsubsection{Problem Statement}

{\color{black}
The use of SLMs and PEFT techniques provides theoretically efficient solutions. However, the challenges associated with implementing and fine-tuning personalized language models on resource-constrained edge devices have yet to be resolved. There are no comprehensive data or standardized metrics available to assist developers in determining the best combination of model architecture (i.e. Transformers vs. SSM) and fine-tuning strategies (LoRA, LoRA+, QLoRA, and BitFit).

As long as there are no quantitative assessments of the trade-offs between model performance (specifically for user-personalized tasks) and operating expenses (such as VRAM usage, time required to train, etc.), it will be hard for companies to effectively expand their on-device AI deployments. Therefore, the problem is the absence of an evaluation framework and a unified metric that balances model performance with sustainable energy consumption across modern architectures.

{\color{black}
\subsection{Motivation for Model and Task Selections}
We focused on SLMs that range from 1B to 2B parameters based on our used hardware, which has 24 GB of VRAM. Under this setup, a 2B-parameter model represents the upper limit for full fine-tuning. Additionally, we selected models from two distinct architectural families to examine their trade-offs in on-device deployment and fine-tuning:
\begin{itemize}
    \item \textit{Transformer-based (TinyLlama-1.1B, Qwen3-1.7B):} TinyLlama serves as a widely adopted, compact, dense baseline for SLMs, while Qwen3 represents the state-of-the-art SLM within this parameter range.
    \item \textit{SSM-based (Mamba-1.4B, Mamba-2-1.3B):} Structured State Space Models are among the most promising alternatives to Transformer based models, since they offer linear-time sequence processing and a constant memory footprint during inference. Evaluating such models against traditional Transformers is important to determine if the theoretical advantages of SSM architectures apply in resource-constrained environments.
\end{itemize}


We evaluated every configuration on six tasks drawn from two complementary benchmark suites: three GLUE tasks for general language understanding and three LaMP tasks for user personalization. This pairing lets us verify that a PEFT method preserves the core NLP skills of a pre-trained SLM while also adapting it to an individual user's profile, the two capabilities that on-device fine-tuning must deliver at once. The six tasks further cover binary classification, multi-class classification, and regression, so the comparison across architectures and PEFT methods is not tied to a single task format.
\begin{itemize}
    \item \textit{General Understanding (GLUE):} We selected SST-2 (sentiment analysis), QNLI (question-answering natural language inference) and STS-B (semantic textual similarity) as a standardized baseline. This ensures that the applied PEFT methods do not cause the models to suffer catastrophic forgetting or lose core NLP skills.
    \item \textit{User Personalization (LaMP):} Because one of the primary motivations of using on-device fine-tuning is to preserve privacy, we used benchmark tasks related to personalization, particularly we selected LaMP-1 (personalized citation identification), LaMP-2 (personalized movie tagging), and LaMP-3 (personalized product rating). These tasks simulate real-world scenarios in which a model must adjust its responses based on a specific user's historical profile.
\end{itemize}

}
\textbf{Research Objectives}: 
To address the gap identified in previous studies regarding the evaluation of on-device AI personalization and deployment efficiencies, this research is driven by the following questions:
\begin{enumerate}
    \item \textbf{Efficiency-Accuracy Trade-off:} To what degree are PEFT methods able to produce results comparable to those obtained through full fine-tuning of SLMs on GLUE and LaMP benchmarks?
    
    \item \textbf{Architecture Comparison:} What differences are there in regards to performance on tasks and computational efficiency between transformer-based models (TinyLlama-1.1B and Qwen3-1.7B) and state-space models (Mamba-1.4B and Mamba-2-1.3B) when fine-tuned using the same PEFT methods?
    
    \item \textbf{Method Selection for Constraints:} Using the NetScore metric and its extensions, what specific combinations of PEFT approaches and model architectures provide the optimal balance between task accuracy and low energy consumption for resource-constrained deployments?
    
\end{enumerate}
}

\subsection{Related Work}



\noindent\textbf{Parameter-Efficient Fine-Tuning}.
The full fine-tuning performs complete update of all the weights of a pre-trained model; therefore, a significant amount of memory is needed to store the entire model copy. 
This adds a need for more efficient methods of adapting pre-trained models. The Parameter Efficient Fine Tuning (PEFT) method~\cite{pu2023empiricalanalysisstrengthsweaknesses} is used to address this problem. In PEFT, the majority of the weights of a pre-trained model are frozen and only a few of the weights are allowed to be adapted during fine-tuning. As a result, a relatively small number of weights needs to be stored and modified. There are several variants of the PEFT method including LoRA~\cite{hu2021loralowrankadaptationlarge}, LoRA+~\cite{hayou2024loraefficientlowrank} and QLoRA~\cite{dettmers2023qloraefficientfinetuningquantized}. The LoRA variant represents each weight update $\Delta W$ as a product of two low-rank matrices $BA$. These two matrices account for only a small fraction of the total number of parameters. 
In addition, the LoRa+ variant introduces different learning rates for the two matrices $A$ and $B$. This leads to better convergence speed and stability. The QLoRA variant uses 4-bit quantization for the weights to further reduce the memory usage. Finally, the BitFit\cite{benzaken2022bitfitsimpleparameterefficientfinetuning} method freezes all the weights except biases, offering the smallest trainable parameter count at the cost of limited performance. However, none of these methods was evaluated against the others under a common framework based on a single set of small models, which is what we address here.

\noindent\textbf{Small Language Models (SLMs)}.
Another approach to efficiency is to start from a small but competent model rather than adapt a large one. Wang et al.~\cite{wang2024comprehensivesurveysmalllanguage} define an SLM by two bounds: the smallest size that still shows emergent ability on a specialized task and the largest size that remains sustainable under the available resource constraints. Their survey covers SLM architectures (Transformers, Mamba, Hymba, xLSTM), methods for deriving SLMs from LLMs (pruning, knowledge distillation, quantization), and the PEFT methods commonly applied to them. Within this range, TinyLlama-1.1B~\cite{zhang2024tinyllamaopensourcesmalllanguage} shows that 1.1 billion parameters trained on three trillion tokens can match or surpass considerably larger models trained on less data, challenging earlier scaling assumptions. Qwen3-1.7B~\cite{yang2025qwen3technicalreport} extends this with roughly 36 trillion tokens across 119 languages, GQA to reduce the key-value cache footprint, and a hybrid thinking mode that trades inference cost for output quality on a per-query basis, reaching performance comparable to Qwen2.5-3B at about half the parameter count. On the deployment side, MobileLLM~\cite{liu2024mobilellmoptimizingsubbillionparameter} combines hardware-friendly architecture changes with post-training quantization to run sub-billion models on mobile hardware, and Lu et al.~\cite{inproceedings} benchmark over 60 SLMs on edge devices, reporting that they can outperform 7B models on general tasks with limited in-context learning.

These works establish SLMs as viable for on-device deployment, but they stop at inference. Neither the survey nor the edge benchmark empirically compares PEFT methods across architecturally distinct models under an energy-aware metric during fine-tuning, which is the gap this work addresses.

\noindent\textbf{State Space Models (SSMs) based SLMs}.
The Mamba~\cite{gu2024mambalineartimesequencemodeling} approach substitutes quadratic self-attention from transformers with selective structured SSMs, whose input-dependent gating provides content-aware reasoning and maintains linear time-complexity. At 1.4 B scale, Mamba produces equivalent language modeling quality as transformers of double size at five times faster throughput.
To improve over the original design, Mamba-2~\cite{dao2024transformersssmsgeneralizedmodels} replaces sequential scanning operation with matrix multiplication, leading to 2-8x faster computation than Mamba while keeping competitive performance.

\section{Methodology}



\subsection{Methodology Overview}

 {\color{black}
As illustrated in Figure~\ref{fig:method}, our methodology pipeline consists of five sequential stages (Benchmarks, Models, PEFT Methods, Performance Metrics, and Comparison):

\begin{figure}[t]
\centering
\includegraphics[width=\columnwidth]{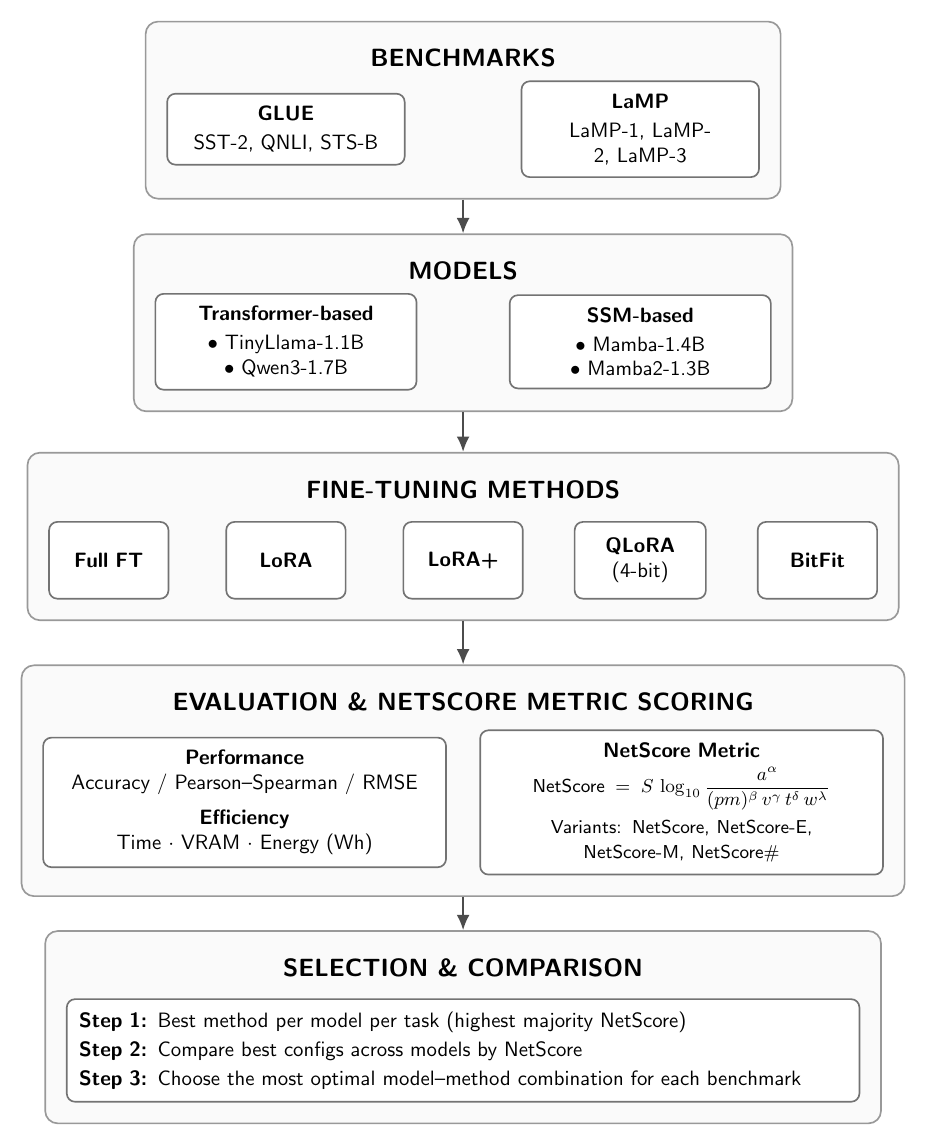}
\caption{Methodology Overview.}
\label{fig:method}
\end{figure}

\textbf{Benchmarks.}
We select six tasks from two different benchmarks. 
Three of the six tasks are selected from each category of the popular GLUE (General Language Understanding Evaluation)~\cite{wang2019gluemultitaskbenchmarkanalysis} benchmark: SST-2 (sentiment analysis), QNLI (natural language inference), and STS-B (Semantic Textual Similarity Benchmark), which are used to assess general-purpose language understanding. 
The remaining three tasks come from the simple and representative LaMP (Language Models Personalization)\cite{salemi2024lamplargelanguagemodels} benchmark: LaMP-1 (Personalized Citation Identification), LaMP-2 (Personalized Movie Tagging), and LaMP-3 (Personalized Product Rating), to evaluate the model's capacity for user-level personalization, which requires fine-tuning strategies.

\textbf{Models.}
Four SLMs are evaluated from two architectural families. First, on the Transformer side, we select TinyLlama-1.1B and Qwen3-1.7B. On the state space model (SSM) side, we choose Mamba-1.4B and Mamba2-1.3B. TinyLlama-1.1B was chosen because it represents a highly optimized, compact, dense transformer trained on an extensive dataset (3 trillion tokens), while Qwen3-1.7B represents the state-of-the-art sub-2B-parameter transformer\cite{yang2025qwen3technicalreport}. SSM-based models were chosen to evaluate their theoretical improved computational and memory efficiency over Transformer models\cite{gu2024mambalineartimesequencemodeling}.

\textbf{Fine-Tuning Methods.}
Each model is adapted to every benchmark task using five fine-tuning methods: Full Fine-Tuning (Full-FT), which updates all parameters, and four PEFT methods, which are LoRA, LoRA+, QLoRA (4-bit quantization), and BitFit. This results in 18 configurations per task (4 models $\times$ 5 methods, minus the two QLoRA runs that the available libraries do not support for the SSM models). 
BitFit was chosen as a lower bound of trainable parameters which represents a minimal baseline for PEFT method. LoRA was chosen as a standard baseline for a PEFT method due to its popularity and widespread use, while LoRA+ was chosen as its improvement with the same parameter budget but faster theoretical convergence and fine-tuning stability. QLoRA was chosen to measure the trade-off between saved VRAM and the potential energy/time costs associated with de-quantization overhead during training.\cite{dettmers2023qloraefficientfinetuningquantized} 

\textbf{Evaluation and Metrics.}
Every configuration is evaluated along two axes. Task performance is measured by accuracy for most tasks (SST-2, QNLI, LaMP-1, LaMP-2), Pearson-Spearman correlation  for STS-B, and Root Mean Squared Error (RMSE) for LaMP-3. Note that, since some tasks have performance metrics where higher scores indicates poorer performance (i.e., errors such as RMSE and MAE), we transform their respective error values prior to generating the NetScore. Specifically,
$\text{perf} = {1}/{(1+\text{error})}$~\cite{mirzal2012pidparametersoptimizationusing}, which is expressed as a percentage (i.e., $a=100/(1+\text{error})$) in the NetScore computation, consistent with the percentage accuracies used for the other tasks.
Resource efficiency is captured through training time, peak VRAM usage, and total energy consumption (Wh). The total energy consumption is obtained through multiplying total training time in hours and average power draw in Watts during training. 

In order to provide a single interpretable value that jointly represents performance and efficiency, we adopt the NetScore metric~\cite{wong2018netscoreuniversalmetricslargescale, wong2019attonetscompactefficientdeep}, originally proposed to summarize the trade-off between accuracy and model size, and extend it to additionally penalize variables related to efficiency. The metric is defined as
$$
\text{NetScore} = S\,\log_{10}\!\left(\frac{a^{\alpha}}{(p\,m)^{\beta}\; v^{\gamma}\; t^{\delta}\; w^{\lambda}}\right),
$$
where $a$ denotes task performance, $p\,m$ the product of the number of parameters $p$ and the FLOPs $m$, $v$ the peak VRAM, $t$ the fine-tuning (or inference) time, and $w$ the average power draw. The scale $S$ and the performance exponent $\alpha$ control how much emphasis is placed on performance relative to cost. Following common practice~\cite{wong2018netscoreuniversalmetricslargescale, wong2019attonetscompactefficientdeep}, we set $S=20$ and $\alpha=2$. The efficiency exponents $\beta,\gamma,\delta,\lambda$ act as switches that determine which cost variables are included, and we use a $1/8$ exponent for the efficiency terms (rather than the $1/4$ used in prior NetScore extensions) to give higher weight to the task performance. 

Rather than a single value, we report a family of NetScore variants per configuration, each prioritizing a different efficiency variable by toggling the corresponding coefficient (Table~\ref{tab:netscore_coeff}): the base \textbf{NetScore}~\cite{wong2018netscoreuniversalmetricslargescale, wong2019attonetscompactefficientdeep} penalizes only model size (parameters and FLOPs); \textbf{NetScore-E}\cite{trinci2024greencontinuallearningreally,wang2026attentionbasedfeaturememorydesign} (energy) penalizes only energy and time; \textbf{NetScore-M}~\cite{toktassyn2026efficient} (memory) penalizes only peak memory; and \textbf{NetScore\#}~\cite{toktassyn2026efficient} combines all efficiency terms. This lets us examine sustainability under several priority levels.


\begin{table}[t]
\centering
\caption{Coefficient settings ($\alpha=2$, $S=20$) defining the NetScore variants. Non-zero efficiency exponents use $1/8 = 0.125$.}
\label{tab:netscore_coeff}
\begin{tabular}{lcccc}
\toprule
\textbf{Variant} & $\beta$ (P$\times$FLOPs) & $\gamma$ (VRAM) & $\delta$ (Time) & $\lambda$ (Power) \\
\midrule
NS~\cite{wong2018netscoreuniversalmetricslargescale, wong2019attonetscompactefficientdeep}    & 0.5 & 0     & 0     & 0     \\
NS-E~\cite{trinci2024greencontinuallearningreally,wang2026attentionbasedfeaturememorydesign}  & 0   & 0     & 0.125 & 0.125 \\
NS-M~\cite{toktassyn2026efficient}  & 0   & 0.125 & 0     & 0     \\
NS\#~\cite{toktassyn2026efficient}  & 0   & 0.125 & 0.125 & 0.125 \\
\bottomrule
\end{tabular}
\end{table}

\textbf{Selection and Comparison.}
The final stage uses NetScores to identify the optimal configuration in three steps: 
(1) \textit{Best PEFT Method across Models}: for each model and task, select the fine-tuning method with the highest energy-focused NS-E, breaking ties on NS-E with the higher NS\#;  
(2) \textit{Best Model across PEFT Methods}: compare the best model across all PEFT methods for each benchmark; and 
(3) \textit{Best PEFT$-$Model Combination across Benchmarks}: determine the most optimal PEFT$-$model combination for each benchmark. 

\section{Experimental Setup}

 
\subsection{Hardware and Software Configuration}
\label{hardware-software-config} 

{\color{black}
As shown in Table~\ref{tab:experimental-setup}, the hardware and software configuration with specific library versions are detailed. All of the experiments were conducted using the same single NVIDIA RTX 4090 (with 24GB of VRAM) and the same software configuration to ensure that variations in train times, VRAM use, and GPU utilization are solely due to the model architecture and the methods used for fine-tuning as opposed to other variables such as environmental variation. Library versions are provided as even small changes to libraries may significantly affect training performance, and therefore impact metrics related to training efficiency.


\begin{table}[ht]
\centering
\caption{Experimental setup.}
\label{tab:experimental-setup}
\begin{tabular}{@{}lp{5.5cm}@{}}
\toprule
\textbf{Component} & \textbf{Specification} \\
\midrule
\multicolumn{2}{l}{\textit{Hardware}} \\
Operating System & Linux (Ubuntu 22.04) \\
GPU & NVIDIA RTX 4090 (24\,GB VRAM) \\
CPU / RAM & Intel Core i9-13900 / 64\,GB DDR5 \\
\midrule
\multicolumn{2}{l}{\textit{Core Software}} \\
Python & 3.11 \\
PyTorch & 2.9.1 (CUDA 12.8, cuDNN 8.9) \\
Hugging Face Transformers & 4.36 (model loading \& tokenisation) \\
Datasets & 2.14 (data pipeline \& preprocessing) \\
PEFT & 0.7 (LoRA / QLoRA adapter training) \\
Evaluate & 0.4 (metric computation) \\
\midrule
\multicolumn{2}{l}{\textit{Quantization}} \\
BitsAndBytes & 0.41 (4-bit NF4 quantization) \\
\midrule
\multicolumn{2}{l}{\textit{SSM-specific Libraries}} \\
causal-conv1d & 1.4.0 (depthwise causal convolution kernel) \\
mamba-ssm & 2.30 (selective-scan CUDA kernel) \\
\midrule
\multicolumn{2}{l}{\textit{Monitoring}} \\
pynvml & 11.5 (GPU utilisation \& VRAM sampled at 100\,ms) \\
\bottomrule
\end{tabular}
\end{table}

\subsection{Models}\label{models} We evaluate four SLMs that belong to two different architectural families. One of them is transformers: TinyLlama-1.1B~\cite{zhang2024tinyllamaopensourcesmalllanguage} (\texttt{TinyLlama/\allowbreak{}TinyLlama-1.1B-\allowbreak{}Chat-v1.0})
 and Qwen3-1.7B~\cite{yang2025qwen3technicalreport} (\texttt{Qwen/Qwen3-1.7B}). 
The SSM-family contains the models Mamba-1.4B~\cite{gu2024mambalineartimesequencemodeling} (\texttt{state-spaces/mamba-1.4b-hf}) and Mamba2-1.3B~\cite{dao2024transformersssmsgeneralizedmodels} (\texttt{state-spaces/mamba2-1.3b}). All models were loaded in bfloat16 precision. 


\subsection{Datasets}\label{datasets}

\subsubsection{GLUE Benchmark}
Three tasks have been selected from the General Language Understanding Evaluation (GLUE) benchmark~\cite{wang2019gluemultitaskbenchmarkanalysis}. These tasks have been directly loaded from Hugging Face Datasets: 
1) SST-2: Stanford Sentiment Treebank (binary sentiment classification, evaluation metric: accuracy).
2) QNLI: Question NLI derived from SQuAD\cite{rajpurkar2016squad100000questionsmachine} (entailment vs. not entailment, evaluation metric: accuracy).
3) STS-B: Semantic Textual Similarity Benchmark (evaluation of regression on similarity scores 0 -- 5, evaluation metric: average of Pearson and Spearman correlation coefficients).

\subsubsection{LaMP Benchmark}
Three tasks have been selected from the Language Models Personalization (LaMP) benchmark~\cite{salemi2024lamplargelanguagemodels} to assess user-level personalization:
1) LaMP-1: Personalized Citation Identification (classification problem, evaluation metric: accuracy).
2) LaMP-2: Personalized Movie Tagging (classification problem; evaluation metric: accuracy).
3) LaMP-3: Personalized Product Rating (regression problem, evaluation metric: Root Mean Squared Error (RMSE)).
To retrieve user-profile information for each LaMP task, we use Contriever~\cite{izacard2022unsuperviseddenseinformationretrieval} (\texttt{facebook/contriever}) with $k=4$ retrieved profile entries per query.

\subsection{Methods for Fine-Tuning}
\label{fine-tune-methods} 
Each model are fine-tuned using five methods:
1) Full Fine-Tuning (Full-FT): All model parameters will be updated during training.
2) LoRA ~\cite{hu2021loralowrankadaptationlarge}: Low-Rank Adapters will be inserted in the attention projections (\texttt{q\_proj}, \texttt{k\_proj}, \texttt{v\_proj}, \texttt{o\_proj}) for transformer-based models, a configuration that showed better performance than the standard query-and-value-only configuration\cite{dettmers2023qloraefficientfinetuningquantized}. The low-rank adapters will be added to \texttt{in\_proj} and \texttt{x\_proj} for Mamba-based models following best practices for SSM fine-tuning\cite{galim2025parameterefficientfinetuningstatespace}. For both architectures, we apply  a rank $r = 16$, a scaling factor $\alpha = 32$, and a dropout of $0.1$. 
3) LoRA+\cite{hayou2024loraefficientlowrank}: The same adapter configuration as LoRA will be applied. However, in contrast to LoRA, this method uses a differentiated learning rate for the $A$ and $B$ matrices. Specifically, a learning rate ratio of $16\times$ will be used (i.e., lr$_B = 16 \cdot$ lr$_A$), which results in faster convergence speed and improved stability\cite{hayou2024loraefficientlowrank}.
4) QLoRA~\cite{dettmers2023qloraefficientfinetuningquantized}: The base model will first be quantized to NF4 precision with double quantization. Afterward, the same adapter configuration as LoRA and LoRA+ will be applied on top.
5) BitFit~\cite{benzaken2022bitfitsimpleparameterefficientfinetuning}: Only the bias terms, layer normalization weights, and the classification heads of the model will be trainable. All other model weights remain frozen.

\subsection{Hyperparameter settings}\label{training-settings} 

Table~\ref{tab:hyperparams} shows the hyperparameter settings for the training of both benchmark tasks. 
For the GLUE experiments, models are trained as sequence classification models with the Hugging Face Trainer\cite{wolf2020huggingfacestransformersstateoftheartnatural} and a linear learning rate schedule with warm-up~\cite{devlin2019bertpretrainingdeepbidirectional}.
A fixed random seed of 42 has been used for all training runs for reproducibility purposes.
For the GLUE experiments, model selection is done based on the best validation metric over every 200 training steps. 

\begin{table}[H]
\centering
\caption{Training Hyperparameters}
\label{tab:hyperparams}
\begin{tabular}{lcc}
\hline
\textbf{Hyperparameter} & \textbf{GLUE} & \textbf{LaMP} \\
\hline
Batch size (per device)       & 32   & 4    \\
Gradient accumulation steps   & 1    & 4    \\
Effective batch size          & 32   & 16   \\
Learning rate                 & $1{\times}10^{-5}$ & $2{\times}10^{-4}$ \\
Epochs                        & 5    & 10   \\
Max sequence length           & 128  & 512  \\
Warmup ratio                  & 0.1  & 0.05 \\
Weight decay                  & 0.01 & 0.01 \\
Precision                     & bf16 & bf16 \\
LoRA rank ($r$)               & 16   & 16   \\
LoRA alpha ($\alpha$)         & 32   & 32   \\
LoRA dropout                  & 0.1  & 0.1  \\
LoRA+ ratio                   & 16   & 16   \\
Random seed                   & 42   & 42   \\
\hline
\end{tabular}
\end{table}

For the LaMP experiments, models are trained as causal language models using generative fine-tuning with beam search ($\text{num\_beams}=4$) at test time, following the evaluation protocol of~\cite{salemi2024lamplargelanguagemodels}.
The hyperparameter and training settings for the LaMP experiments were taken from the original LaMP paper~\cite{salemi2024lamplargelanguagemodels}.  
For the LaMP experiments, up to 128 new tokens can be generated with beam search decoding with 4 beams.
}

\begin{table*}[h!]
\centering
\caption{
Merged results on \textbf{GLUE (SST-2, QNLI, STS-B)} benchmark
(finetuning + inference). ``Perf.'' denotes accuracy for SST-2 and
QNLI and the Pearson--Spearman correlation for STS-B; \textbf{Common}
columns are identical for finetuning and inference. NS = NetScore:
NS-E uses time and power; NS-M uses VRAM; NS\# uses VRAM, time, and
power, all without the parameter/FLOPs term. All variants use the
$1/8$ coefficient. Finetuning Time/Energy are in min/Wh; Inference
Time/Energy are in s/mWh. QLoRA results for Mamba and Mamba2 are
omitted because currently available libraries do not support their
native quantization. Bold numeric values mark the best value within each model group,
computed separately for finetuning and inference. For finetuning,
red highlights the highest NS-E and blue highlights the highest NS-M.
The selected finetuning method (bold, red-shaded method name) follows
a strict energy-first rule: the method with the highest NS-E is
selected, and a tie on NS-E is broken by the higher NS\#. This is
the same rule used in Table~\ref{tab:glue_summary_merged}.
}
\label{tab:sst2_merged}
\resizebox{\textwidth}{!}{%
\begin{tabular}{cll cc ccccccccc ccccccccc}
\toprule
\multirow{3}{*}{\textbf{Task}} &
\multirow{3}{*}{\textbf{Model}} &
\multirow{3}{*}{\textbf{Method}} &
\multicolumn{2}{c}{\textbf{Common}} &
\multicolumn{9}{c}{\textbf{Finetuning}} &
\multicolumn{9}{c}{\textbf{Inference}} \\
\cmidrule(lr){4-5}
\cmidrule(lr){6-14}
\cmidrule(lr){15-23}
& & &
\textbf{Params $\downarrow$} & \textbf{Perf. $\uparrow$} &
\textbf{TFLOPs $\downarrow$} & \textbf{Time $\downarrow$} &
\textbf{VRAM $\downarrow$} & \textbf{Power $\downarrow$} &
\textbf{Energy $\downarrow$} & \textbf{NS $\uparrow$} &
\textbf{NS-E $\uparrow$} & \textbf{NS-M $\uparrow$} &
\textbf{NS\# $\uparrow$} &
\textbf{TFLOPs $\downarrow$} & \textbf{Time $\downarrow$} &
\textbf{VRAM $\downarrow$} & \textbf{Power $\downarrow$} &
\textbf{Energy $\downarrow$} & \textbf{NS $\uparrow$} &
\textbf{NS-E $\uparrow$} & \textbf{NS-M $\uparrow$} &
\textbf{NS\# $\uparrow$} \\
& & & (M) & &
(TF) & (min) & (MB) & (W) & (Wh) & & & & &
(TF) & (s) & (MB) & (W) & (mWh) & & & & \\
\midrule
\multirow{18}{*}{\textbf{SST-2}}
& \multirow{5}{*}{TinyLlama-1.1B}
& Full-FT & 1104.55 & 0.954 & 24.09 & 34.01 & 22850.4 & 314.50 & 178.27 & 34.9 & 69.1 & 68.3 & 58.2 & 8.03 & 15.00 & 3876.0 & \textbf{90.81} & \textbf{378.38} & 39.7 & \textbf{71.3} & 70.2 & \textbf{62.4} \\
& & \cellcolor{red!15}\textbf{LoRA} & 4.51 & \textbf{0.959} & 16.17 & 19.18 & 16479.8 & \textbf{261.43} & \textbf{83.57} & 60.6 & \cellcolor{red!15}\textbf{70.0} & 68.7 & 59.5 & 8.03 & 18.73 & 3654.0 & 94.00 & 489.06 & 63.7 & 71.2 & 70.4 & 62.3 \\
& & LoRA+ & 4.51 & 0.957 & 16.17 & 19.33 & 15957.0 & 268.84 & 86.61 & 60.6 & 69.9 & 68.7 & 59.4 & 8.03 & 16.65 & 3765.0 & 94.50 & 437.06 & 63.6 & 71.2 & 70.3 & 62.3 \\
& & QLoRA & 4.51 & 0.952 & 16.17 & 31.05 & \textbf{4225.2} & 284.56 & 147.26 & 60.5 & 69.3 & \cellcolor{blue!15}\textbf{70.1} & \textbf{60.2} & 8.03 & 29.36 & \textbf{3243.0} & 116.70 & 951.75 & 63.6 & 70.3 & \textbf{70.4} & 61.5 \\
& & BitFit & \textbf{0.09} & 0.920 & \textbf{16.06} & \textbf{18.85} & 17950.0 & 288.41 & 90.61 & \textbf{77.0} & 69.2 & 67.9 & 58.6 & 8.03 & \textbf{14.86} & 3608.0 & 91.76 & 378.76 & \textbf{80.0} & 70.7 & 69.7 & 61.8 \\
\cmidrule(lr){2-23}
& \multirow{5}{*}{Qwen3-1.7B}
& Full-FT & 1720.57 & 0.955 & 35.00 & 51.56 & 22586.7 & 338.59 & 290.96 & 31.4 & 68.6 & 68.3 & 57.7 & 11.67 & \textbf{17.00} & 4812.0 & \textbf{98.47} & \textbf{465.00} & 36.2 & \textbf{71.1} & 70.0 & \textbf{61.9} \\
& & LoRA & 6.42 & 0.955 & 23.49 & \textbf{21.46} & 23977.5 & 360.31 & 128.87 & 57.4 & 69.5 & 68.3 & 58.5 & 11.67 & 23.00 & 4467.0 & 102.00 & 651.67 & 60.5 & 70.8 & 70.1 & 61.6 \\
& & \cellcolor{red!15}\textbf{LoRA+} & 6.42 & \textbf{0.962} & 23.49 & 22.30 & 20047.6 & 350.26 & 130.18 & 57.5 & \cellcolor{red!15}\textbf{69.6} & 68.6 & 58.8 & 11.67 & 23.00 & 4678.0 & 101.00 & 645.28 & 60.6 & 70.9 & 70.2 & 61.7 \\
& & QLoRA & 6.42 & 0.958 & 23.49 & 34.17 & \textbf{7808.1} & 376.07 & 214.17 & 57.5 & 69.0 & \cellcolor{blue!15}\textbf{69.5} & \textbf{59.3} & 11.67 & 34.00 & \textbf{3120.0} & 138.00 & 1303.33 & 60.5 & 70.1 & \textbf{70.5} & 61.3 \\
& & BitFit & \textbf{0.12} & 0.900 & \textbf{23.33} & 22.27 & 18683.6 & \textbf{333.06} & \textbf{123.62} & \textbf{73.7} & 68.5 & 67.5 & 57.8 & 11.67 & \textbf{17.00} & 4884.0 & 99.00 & 467.50 & \textbf{76.7} & 70.1 & 68.9 & 60.9 \\
\cmidrule(lr){2-23}
& \multirow{4}{*}{Mamba-1.4B}
& Full-FT & 1383.31 & \textbf{0.958} & 31.45 & 71.41 & 24419.7 & \textbf{379.54} & 451.71 & 32.9 & 68.2 & 68.3 & 57.2 & 10.48 & \textbf{17.00} & 4151.0 & 99.00 & 467.50 & 37.6 & \textbf{71.2} & 70.2 & \textbf{62.1} \\
& & LoRA & 11.13 & 0.957 & 21.24 & 48.71 & 10795.8 & 387.37 & 314.48 & 55.5 & 68.5 & \cellcolor{blue!15}\textbf{69.2} & 58.5 & 10.48 & 23.50 & 4012.0 & 101.00 & 659.31 & 58.6 & 70.8 & 70.2 & 61.8 \\
& & \cellcolor{red!15}\textbf{LoRA+} & 11.13 & \textbf{0.958} & 21.24 & 49.37 & \textbf{10779.6} & 383.87 & 315.86 & 55.5 & \cellcolor{red!15}\textbf{68.6} & \cellcolor{blue!15}\textbf{69.2} & \textbf{58.5} & 10.48 & 23.50 & 3970.0 & 105.00 & 685.42 & 58.6 & 70.8 & \textbf{70.3} & 61.8 \\
& & BitFit & \textbf{0.39} & 0.809 & \textbf{20.96} & \textbf{44.93} & 22646.9 & 394.41 & \textbf{295.35} & \textbf{67.2} & 65.7 & 65.4 & 54.8 & 10.48 & \textbf{17.00} & \textbf{3852.0} & \textbf{98.00} & \textbf{462.78} & \textbf{70.2} & 68.3 & 67.4 & 59.3 \\
\cmidrule(lr){2-23}
& \multirow{4}{*}{Mamba2-1.3B}
& Full-FT & 1351.87 & 0.936 & 33.26 & 46.16 & 12990.9 & \textbf{330.82} & 254.51 & 32.3 & 68.4 & 68.6 & 58.1 & 11.09 & \textbf{29.00} & \textbf{4190.0} & \textbf{98.00} & \textbf{789.44} & 37.1 & 70.2 & 69.8 & 61.2 \\
& & LoRA & 8.11 & 0.955 & 22.38 & 34.24 & 12903.4 & 338.00 & 192.88 & 56.6 & 69.0 & 68.9 & 58.8 & 11.09 & 31.80 & 4225.0 & 104.00 & 918.67 & 59.7 & 70.4 & 70.1 & 61.3 \\
& & \cellcolor{red!15}\textbf{LoRA+} & 8.11 & \textbf{0.967} & 22.38 & \textbf{31.70} & 12805.1 & 357.50 & 188.88 & 56.8 & \cellcolor{red!15}\textbf{69.3} & \cellcolor{blue!15}\textbf{69.1} & \textbf{59.0} & 11.09 & 31.70 & 4356.0 & 104.00 & 915.78 & 59.9 & \textbf{70.6} & \textbf{70.3} & \textbf{61.5} \\
& & BitFit & \textbf{0.21} & 0.500 & \textbf{22.18} & 32.84 & \textbf{9029.5} & 331.83 & \textbf{181.62} & \textbf{61.3} & 57.9 & 58.1 & 48.0 & 11.09 & \textbf{29.00} & 4368.0 & 101.00 & 813.61 & \textbf{64.3} & 59.3 & 58.9 & 50.2 \\
\midrule
\midrule
\multirow{18}{*}{\textbf{QNLI}}
& \multirow{5}{*}{TinyLlama-1.1B}
& Full-FT & 1104.55 & \textbf{0.934} & 24.09 & 109.33 & 23200.1 & 325.32 & 592.79 & 34.6 & 67.4 & 67.9 & 56.5 & 8.03 & \textbf{141.00} & 3768.0 & 85.16 & 3335.43 & 39.3 & \textbf{68.6} & 69.9 & 59.7 \\
& & LoRA & 4.51 & 0.924 & 16.17 & 78.13 & 16699.1 & 280.30 & \textbf{364.99} & 60.0 & 67.8 & 68.1 & 57.2 & 8.03 & 153.00 & 3168.0 & 87.27 & 3708.98 & 63.0 & 68.3 & 69.9 & 59.6 \\
& & \cellcolor{red!15}\textbf{LoRA+} & 4.51 & 0.932 & 16.17 & 81.46 & 16728.6 & 283.19 & 384.49 & 60.1 & \cellcolor{red!15}\textbf{67.9} & 68.2 & 57.3 & 8.03 & 153.00 & 3158.0 & 86.53 & 3677.53 & 63.2 & 68.5 & 70.0 & \textbf{59.7} \\
& & QLoRA & 4.51 & 0.917 & 16.17 & 124.59 & \textbf{4797.4} & \textbf{273.32} & 567.55 & 59.9 & 67.2 & \cellcolor{blue!15}\textbf{69.3} & \textbf{58.0} & 8.03 & 184.00 & \textbf{2120.0} & 110.32 & 5638.58 & 62.9 & 67.7 & \textbf{70.2} & 59.4 \\
& & BitFit & \textbf{0.09} & 0.839 & \textbf{16.06} & \textbf{75.07} & 19450.2 & 308.12 & 385.51 & \textbf{75.4} & 66.0 & 66.2 & 55.3 & 8.03 & \textbf{141.00} & 3132.0 & \textbf{84.82} & \textbf{3322.12} & \textbf{78.4} & 66.8 & 68.2 & 58.0 \\
\cmidrule(lr){2-23}
& \multirow{5}{*}{Qwen3-1.7B}
& Full-FT & 1720.57 & 0.942 & 35.00 & 170.30 & 24089.7 & 390.01 & 1106.99 & 31.2 & 66.9 & 68.0 & 56.0 & 11.67 & 153.00 & 5678.0 & \textbf{94.23} & 4004.78 & 35.9 & 68.6 & 69.6 & 59.2 \\
& & LoRA & 6.42 & 0.943 & 23.49 & 102.80 & 23674.0 & 395.16 & 677.04 & 57.2 & \cellcolor{red!15}\textbf{67.5} & 68.0 & 56.5 & 11.67 & 166.00 & 5684.0 & 97.57 & 4499.06 & 60.2 & 68.5 & 69.6 & 59.1 \\
& & \cellcolor{red!15}\textbf{LoRA+} & 6.42 & \textbf{0.946} & 23.49 & 104.20 & 23605.8 & 391.14 & 679.28 & 57.3 & \cellcolor{red!15}\textbf{67.5} & 68.1 & 56.6 & 11.67 & \textbf{144.00} & 5668.0 & 99.50 & \textbf{3980.00} & 60.3 & \textbf{68.6} & 69.7 & \textbf{59.3} \\
& & QLoRA & 6.42 & 0.939 & 23.49 & 159.40 & \textbf{11114.3} & \textbf{387.68} & 1029.93 & 57.1 & 66.9 & \cellcolor{blue!15}\textbf{68.8} & \textbf{56.8} & 11.67 & 213.00 & \textbf{4040.0} & 135.59 & 8022.41 & 60.2 & 67.8 & \textbf{69.9} & 58.7 \\
& & BitFit & \textbf{0.12} & 0.859 & \textbf{23.33} & \textbf{99.40} & 23339.0 & 388.94 & \textbf{644.35} & \textbf{72.9} & 65.9 & 66.4 & 55.0 & 11.67 & 152.00 & 5592.0 & 94.79 & 4002.24 & \textbf{75.9} & 67.0 & 68.0 & 57.6 \\
\cmidrule(lr){2-23}
& \multirow{4}{*}{Mamba-1.4B}
& Full-FT & 1383.31 & 0.902 & 31.45 & 197.66 & 24531.9 & \textbf{413.36} & 1361.73 & 31.8 & 65.9 & 67.2 & 55.0 & 10.48 & \textbf{112.00} & 4120.0 & \textbf{94.43} & \textbf{2937.82} & 36.6 & 68.1 & 69.2 & 59.1 \\
& & LoRA & 11.13 & 0.920 & 21.24 & 147.83 & \textbf{13688.1} & 422.05 & 1039.87 & 54.8 & 66.6 & \cellcolor{blue!15}\textbf{68.2} & \textbf{56.2} & 10.48 & 148.00 & 3986.0 & 97.52 & 4009.16 & 57.9 & 68.2 & 69.6 & 59.2 \\
& & \cellcolor{red!15}\textbf{LoRA+} & 11.13 & \textbf{0.928} & 21.24 & 147.90 & 19639.4 & 421.40 & 1038.76 & 55.0 & \cellcolor{red!15}\textbf{66.7} & 68.0 & 56.0 & 10.48 & 148.00 & 3942.0 & 97.61 & 4012.86 & 58.0 & \textbf{68.3} & \textbf{69.7} & \textbf{59.3} \\
& & BitFit & \textbf{0.39} & 0.675 & \textbf{20.96} & \textbf{132.50} & 19153.5 & 424.45 & \textbf{937.32} & \textbf{64.0} & 61.3 & 62.5 & 50.6 & 10.48 & 153.00 & \textbf{3840.0} & 96.28 & 4091.90 & \textbf{67.1} & 62.8 & 64.2 & 53.8 \\
\cmidrule(lr){2-23}
& \multirow{4}{*}{Mamba2-1.3B}
& Full-FT & 1351.87 & 0.824 & 33.26 & 119.56 & 21769.0 & \textbf{392.03} & 781.18 & 30.1 & 65.0 & 65.8 & 54.1 & 11.09 & \textbf{187.00} & 4567.0 & \textbf{92.77} & \textbf{4818.89} & 34.9 & 66.0 & 67.5 & 56.9 \\
& & LoRA & 8.11 & 0.917 & 22.38 & 100.24 & 17756.4 & 413.83 & 691.36 & 55.9 & 67.0 & 67.9 & 56.3 & 11.09 & 199.00 & 4442.0 & 99.69 & 5510.64 & 59.0 & 67.8 & 69.4 & 58.6 \\
& & \cellcolor{red!15}\textbf{LoRA+} & 8.11 & \textbf{0.926} & 22.38 & 101.35 & 17726.1 & 406.87 & 687.27 & 56.1 & \cellcolor{red!15}\textbf{67.1} & \cellcolor{blue!15}\textbf{68.0} & \textbf{56.5} & 11.09 & 197.00 & 4390.0 & 98.75 & 5403.82 & 59.1 & \textbf{67.9} & \textbf{69.6} & \textbf{58.8} \\
& & BitFit & \textbf{0.21} & 0.500 & \textbf{22.18} & \textbf{91.96} & \textbf{16521.3} & 406.89 & \textbf{623.63} & \textbf{62.4} & 57.6 & 58.5 & 47.1 & 11.09 & \textbf{187.00} & \textbf{4304.0} & 93.73 & 4868.75 & \textbf{65.4} & 58.5 & 60.0 & 49.4 \\
\midrule
\midrule
\multirow{18}{*}{\textbf{STS-B}}
& \multirow{5}{*}{TinyLlama-1.1B}
& Full-FT & 1104.55 & \textbf{0.909} & 24.09 & 4.75 & 12444.4 & 371.24 & 29.39 & 34.1 & 70.2 & \cellcolor{blue!15}\textbf{68.1} & \textbf{60.0} & 8.03 & 22.00 & 5850.0 & 105.59 & 645.27 & 38.9 & \textbf{69.9} & \textbf{68.9} & \textbf{60.5} \\
& & LoRA & 4.51 & 0.839 & 16.17 & 2.50 & 24039.3 & 374.88 & 15.62 & 58.3 & 69.5 & 66.0 & 58.6 & 8.03 & 33.00 & 5574.0 & 102.22 & 937.02 & 61.4 & 68.1 & 67.6 & 58.8 \\
& & \cellcolor{red!15}\textbf{LoRA+} & 4.51 & 0.893 & 16.17 & 2.56 & 19197.1 & 367.73 & 15.69 & \textbf{59.4} & \cellcolor{red!15}\textbf{70.6} & 67.3 & 59.9 & 8.03 & 33.00 & 5676.0 & 102.38 & 938.48 & \textbf{62.4} & 69.2 & 68.6 & 59.8 \\
& & QLoRA & 4.51 & 0.829 & 16.17 & 4.20 & \textbf{8893.5} & 373.29 & 26.13 & 58.1 & 68.8 & 66.9 & 58.9 & 8.03 & 47.00 & \textbf{3350.0} & 118.90 & 1552.31 & 61.2 & 67.4 & 67.9 & 58.6 \\
& & BitFit & \textbf{0.09} & 0.310 & \textbf{16.06} & \textbf{2.44} & 24209.6 & \textbf{357.79} & \textbf{14.55} & 58.1 & 52.3 & 48.7 & 41.3 & 8.03 & \textbf{21.00} & 5144.0 & \textbf{100.28} & \textbf{584.97} & 61.1 & 51.3 & 50.4 & 42.1 \\
\cmidrule(lr){2-23}
& \multirow{5}{*}{Qwen3-1.7B}
& Full-FT & 1720.57 & \textbf{0.903} & 35.00 & 12.63 & 24482.4 & \textbf{228.55} & 48.11 & 30.4 & 69.6 & \cellcolor{blue!15}\textbf{67.3} & 58.6 & 11.67 & \textbf{30.00} & 10293.0 & \textbf{98.50} & \textbf{820.83} & 35.2 & \textbf{69.6} & \textbf{68.2} & \textbf{59.5} \\
& & LoRA & 6.42 & 0.795 & 23.49 & \textbf{3.31} & 23809.7 & 367.07 & 20.25 & 54.2 & 68.3 & 65.1 & 57.4 & 11.67 & 41.00 & 9592.0 & 117.81 & 1341.73 & 57.3 & 66.8 & 66.1 & 56.9 \\
& & \cellcolor{red!15}\textbf{LoRA+} & 6.42 & 0.893 & 23.49 & 3.42 & 23622.6 & 358.60 & 20.44 & \textbf{56.2} & \cellcolor{red!15}\textbf{70.3} & 67.1 & \textbf{59.4} & 11.67 & 41.00 & 9536.0 & 118.15 & 1345.60 & \textbf{59.3} & 68.8 & 68.1 & 58.9 \\
& & QLoRA & 6.42 & 0.818 & 23.49 & 5.41 & \textbf{7545.5} & 370.43 & 33.40 & 54.7 & 68.3 & 66.8 & 58.6 & 11.67 & 58.00 & \textbf{6689.0} & 141.41 & 2278.27 & 57.8 & 66.7 & 66.9 & 57.2 \\
& & BitFit & \textbf{0.12} & 0.316 & \textbf{23.33} & 3.37 & 23435.9 & 345.40 & \textbf{19.40} & 55.5 & 52.3 & 49.1 & 41.4 & 11.67 & 36.00 & 7728.0 & 115.00 & 1150.00 & 58.5 & 50.9 & 50.3 & 41.2 \\
\cmidrule(lr){2-23}
& \multirow{4}{*}{Mamba-1.4B}
& Full-FT & 1383.31 & 0.814 & 31.45 & 7.57 & \textbf{10887.5} & \textbf{408.51} & 51.54 & 30.0 & 67.7 & 66.3 & 57.6 & 10.48 & 32.00 & 4450.0 & 99.37 & 883.29 & 34.8 & 67.7 & 67.3 & 58.5 \\
& & LoRA & 11.13 & 0.807 & 21.24 & 5.12 & 18525.9 & 422.46 & 36.05 & 52.5 & 67.9 & 65.6 & 57.3 & 10.48 & 43.00 & 4120.0 & 102.94 & 1229.56 & 55.6 & 67.2 & 67.2 & 58.1 \\
& & \cellcolor{red!15}\textbf{LoRA+} & 11.13 & \textbf{0.869} & 21.24 & 5.18 & 18798.0 & 418.61 & 36.14 & \textbf{53.8} & \cellcolor{red!15}\textbf{69.2} & \cellcolor{blue!15}\textbf{66.9} & \textbf{58.5} & 10.48 & 43.00 & 4526.0 & 104.00 & 1242.22 & \textbf{56.9} & \textbf{68.4} & \textbf{68.4} & \textbf{59.3} \\
& & BitFit & \textbf{0.39} & 0.146 & \textbf{20.96} & \textbf{4.83} & 19158.6 & 417.14 & \textbf{33.58} & 37.4 & 38.3 & 35.9 & 27.6 & 10.48 & \textbf{31.00} & \textbf{3840.0} & \textbf{98.58} & \textbf{848.88} & 40.5 & 37.9 & 37.6 & 28.9 \\
\cmidrule(lr){2-23}
& \multirow{4}{*}{Mamba2-1.3B}
& Full-FT & 1351.87 & 0.638 & 33.26 & 4.935 & 16998.34 & \textbf{356.23} & 29.30 & 25.7 & 64.1 & 61.6 & 53.5 & 11.09 & 51.00 & 4777.0 & \textbf{98.75} & 1398.96 & 30.4 & 62.9 & 63.0 & 53.7 \\
& & LoRA & 8.11 & 0.705 & 22.38 & \textbf{3.50} & 15044.3 & 384.17 & 22.41 & 51.3 & 66.1 & 63.5 & 55.7 & 11.09 & 53.00 & 4566.0 & 100.89 & 1485.33 & 54.4 & 64.6 & 64.8 & 55.5 \\
& & \cellcolor{red!15}\textbf{LoRA+} & 8.11 & \textbf{0.847} & 22.38 & 3.54 & 14986.6 & 381.36 & 22.50 & \textbf{54.5} & \cellcolor{red!15}\textbf{69.3} & \cellcolor{blue!15}\textbf{66.7} & \textbf{58.9} & 11.09 & 52.00 & 4488.0 & 102.00 & 1473.33 & \textbf{57.6} & \textbf{67.8} & \textbf{68.0} & \textbf{58.7} \\
& & BitFit & \textbf{0.21} & 0.023 & \textbf{22.18} & 3.53 & \textbf{12893.7} & 366.97 & \textbf{21.59} & 7.8 & 6.7 & 4.2 & -3.6 & 11.09 & \textbf{47.00} & \textbf{4368.0} & 99.71 & \textbf{1301.77} & 10.8 & 5.3 & 5.4 & -3.8 \\
\bottomrule
\end{tabular}%
}
\end{table*}

\begin{table*}[h!]
\centering
\caption{
Merged results on \textbf{LaMP (LaMP-1/2/3)} benchmark
(finetuning + inference). ``Perf.'' denotes accuracy for LaMP-1 and
LaMP-2 (higher is better) and RMSE for LaMP-3 (lower is better).
\textbf{Common} columns are identical for finetuning and inference.
NS = NetScore: NS-E uses time and power; NS-M uses VRAM; NS\# uses
VRAM, time, and power, all without the parameter/FLOPs term. All
variants use the $1/8$ coefficient. Finetuning Time/Energy are in
min/Wh; inference Time/Energy are in min/Wh for LaMP-1 and LaMP-3 and
in s/mWh for LaMP-2. QLoRA results for Mamba and Mamba2
are omitted because currently available libraries do not support their
native quantization. Bold numeric values mark the best value within
each model group, computed separately for finetuning and inference.
For finetuning, red highlights the highest NS-E and blue highlights
the highest NS-M. The selected finetuning method (bold, red-shaded
method name) follows a strict energy-first rule: the method with the
highest NS-E is selected, and a tie on NS-E is broken by the higher
NS\#. This is the same rule used in
Table~\ref{tab:lamp_summary_merged}.
}
\label{tab:lamp1_merged}
\resizebox{\textwidth}{!}{%
\begin{tabular}{cll cc ccccccccc ccccccccc}
\toprule
\multirow{3}{*}{\textbf{Task}} &
\multirow{3}{*}{\textbf{Model}} &
\multirow{3}{*}{\textbf{Method}} &
\multicolumn{2}{c}{\textbf{Common}} &
\multicolumn{9}{c}{\textbf{Finetuning}} &
\multicolumn{9}{c}{\textbf{Inference}} \\
\cmidrule(lr){4-5}
\cmidrule(lr){6-14}
\cmidrule(lr){15-23}
& & &
\textbf{Params $\downarrow$} & \textbf{Perf. $\uparrow/\downarrow$} &
\textbf{TFLOPs $\downarrow$} & \textbf{Time $\downarrow$} &
\textbf{VRAM $\downarrow$} & \textbf{Power $\downarrow$} &
\textbf{Energy $\downarrow$} & \textbf{NS $\uparrow$} &
\textbf{NS-E $\uparrow$} & \textbf{NS-M $\uparrow$} &
\textbf{NS\# $\uparrow$} &
\textbf{TFLOPs $\downarrow$} & \textbf{Time $\downarrow$} &
\textbf{VRAM $\downarrow$} & \textbf{Power $\downarrow$} &
\textbf{Energy $\downarrow$} & \textbf{NS $\uparrow$} &
\textbf{NS-E $\uparrow$} & \textbf{NS-M $\uparrow$} &
\textbf{NS\# $\uparrow$} \\
& & & (M) & &
(TF) & (min) & (MB) & (W) & (Wh) & & & & &
(TF) & (min/s) & (MB) & (W) & (Wh/mWh) & & & & \\
\midrule

\multirow{18}{*}{\textbf{LaMP-1}}
& \multirow{5}{*}{TinyLlama-1.1B}
& Full-FT & 1104.55 & 0.707 & 25.99 & 40.9 & 12424.3 & 376.28 & 256.5 & 29.4 & 63.5 & 63.7 & 53.3 & 8.66 & 11.93 & 3944.0 & 323.21 & \textbf{64.28} & 34.2 & 65.0 & 65.0 & 56.0 \\
& & LoRA & 4.51 & 0.730 & 17.44 & 23.4 & 8039.6 & 369.49 & 144.1 & 55.6 & \cellcolor{red!15}\textbf{64.7} & 64.8 & 54.9 & 8.66 & 12.45 & 3966.0 & 314.00 & 65.16 & 58.6 & 65.6 & 65.5 & 56.6 \\
& & \cellcolor{red!15}\textbf{LoRA+} & 4.51 & \textbf{0.732} & 17.44 & 27.5 & 7608.5 & 324.22 & 148.6 & 55.6 & \cellcolor{red!15}\textbf{64.7} & \cellcolor{blue!15}\textbf{64.9} & \textbf{55.0} & 8.66 & 12.52 & 3966.0 & 313.50 & 65.40 & 58.7 & \textbf{65.6} & 65.6 & 56.6 \\
& & QLoRA & 4.51 & 0.728 & 17.44 & 39.8 & 7451.8 & \textbf{282.81} & 187.6 & 55.5 & 64.4 & 64.8 & 54.7 & 8.66 & 13.82 & \textbf{2580.0} & \textbf{292.44} & 67.34 & 58.6 & 65.5 & \textbf{66.0} & \textbf{56.9} \\
& & BitFit & \textbf{0.09} & 0.480 & \textbf{17.33} & \textbf{20.9} & \textbf{7361.7} & 384.11 & \textbf{133.8} & \textbf{65.3} & 57.5 & 57.6 & 47.8 & 8.66 & \textbf{11.73} & 3944.0 & 331.85 & 64.90 & \textbf{68.3} & 58.3 & 58.3 & 49.3 \\
\cmidrule(lr){2-23}

& \multirow{5}{*}{Qwen3-1.7B}
& Full-FT & 1720.57 & 0.739 & 43.00 & 61.2 & 22690.6 & 358.92 & 366.1 & 26.1 & 63.9 & 63.9 & 53.0 & 14.33 & \textbf{12.33} & 5080.0 & 321.00 & \textbf{65.98} & 30.8 & 65.8 & 65.5 & 56.5 \\
& & \cellcolor{red!15}\textbf{LoRA} & 6.42 & \textbf{0.760} & 28.83 & 34.3 & 19097.9 & 367.35 & 210.0 & 52.6 & \cellcolor{red!15}\textbf{65.0} & \cellcolor{blue!15}\textbf{64.5} & \textbf{54.3} & 14.33 & 13.33 & 5458.0 & 305.00 & 67.78 & 55.6 & \textbf{66.2} & \textbf{65.9} & \textbf{56.9} \\
& & LoRA+ & 6.42 & 0.758 & 28.83 & 40.6 & 18992.9 & 324.24 & 219.4 & 52.5 & 64.9 & \cellcolor{blue!15}\textbf{64.5} & 54.2 & 14.33 & 13.22 & 5490.0 & 307.00 & 67.63 & 55.5 & 66.2 & 65.8 & 56.8 \\
& & QLoRA & 6.42 & 0.736 & 28.83 & 54.8 & \textbf{16402.2} & \textbf{288.72} & 263.7 & 52.0 & 64.2 & 64.1 & 53.6 & 14.33 & 15.27 & \textbf{3916.0} & 281.00 & 71.50 & 55.0 & 65.6 & 65.7 & 56.6 \\
& & BitFit & \textbf{0.12} & 0.439 & \textbf{28.67} & \textbf{31.7} & 18392.2 & 382.15 & \textbf{201.9} & \textbf{60.3} & 55.5 & 55.0 & 44.8 & 14.33 & 23.77 & 5014.0 & \textbf{261.00} & 103.39 & \textbf{63.3} & 56.2 & 56.4 & 47.0 \\
\cmidrule(lr){2-23}

& \multirow{4}{*}{Mamba-1.4B}
& Full-FT & 1383.31 & 0.670 & 33.98 & 67.0 & 14600.5 & 392.60 & 438.4 & 26.3 & 62.0 & 62.6 & 51.6 & 11.33 & 30.53 & 4294.0 & 215.00 & 109.41 & 31.1 & 63.5 & 64.0 & 54.4 \\
& & LoRA & 11.13 & 0.570 & 22.92 & 48.0 & 11260.7 & 381.00 & 304.8 & 46.2 & 59.6 & 60.1 & 49.5 & 11.33 & 40.65 & 4326.0 & \textbf{185.00} & 125.34 & 49.2 & 60.5 & 61.1 & 51.5 \\
& & \cellcolor{red!15}\textbf{LoRA+} & 11.13 & \textbf{0.698} & 22.92 & 50.8 & 11124.9 & \textbf{368.74} & 312.2 & 49.7 & \cellcolor{red!15}\textbf{63.1} & \cellcolor{blue!15}\textbf{63.6} & \textbf{53.0} & 11.33 & 36.93 & 4326.0 & 195.00 & 120.03 & 52.7 & \textbf{64.1} & \textbf{64.7} & \textbf{55.0} \\
& & BitFit & \textbf{0.39} & 0.501 & \textbf{22.65} & \textbf{40.6} & \textbf{10701.4} & 417.19 & \textbf{282.3} & \textbf{58.5} & 57.4 & 57.9 & 47.3 & 11.33 & \textbf{29.90} & \textbf{4254.0} & 216.00 & \textbf{107.64} & \textbf{61.5} & 58.5 & 58.9 & 49.4 \\
\cmidrule(lr){2-23}

& \multirow{4}{*}{Mamba2-1.3B}
& Full-FT & 1351.87 & 0.686 & 35.79 & 77.9 & 15290.4 & 230.91 & 299.8 & 26.6 & 62.8 & 63.0 & 52.4 & 11.93 & 31.08 & 4178.0 & 221.00 & 114.49 & 31.4 & 63.9 & 64.4 & 54.8 \\
& & LoRA & 8.11 & 0.700 & 24.06 & \textbf{44.7} & 8013.4 & 286.44 & \textbf{213.4} & 50.9 & 63.5 & 64.0 & 53.8 & 11.93 & 39.17 & 4212.0 & \textbf{189.00} & 123.38 & 53.9 & 64.1 & 64.7 & 55.1 \\
& & \cellcolor{red!15}\textbf{LoRA+} & 8.11 & \textbf{0.722} & 24.06 & 78.0 & 7680.3 & 198.92 & 258.6 & 51.4 & \cellcolor{red!15}\textbf{63.9} & \cellcolor{blue!15}\textbf{64.6} & \textbf{54.2} & 11.93 & 38.33 & \textbf{4145.0} & 190.00 & 121.39 & 54.5 & \textbf{64.7} & \textbf{65.3} & \textbf{55.6} \\
& & BitFit & \textbf{0.21} & 0.517 & \textbf{23.86} & 69.8 & \textbf{7131.7} & \textbf{187.31} & 217.9 & \textbf{61.5} & 58.2 & 58.9 & 48.6 & 11.93 & \textbf{29.08} & 4200.0 & 217.00 & \textbf{105.18} & \textbf{64.6} & 59.0 & 59.5 & 50.0 \\

\midrule
\midrule

\multirow{18}{*}{\textbf{LaMP-2}}
& \multirow{5}{*}{TinyLlama-1.1B}
& Full-FT & 1104.55 & 0.696 & 25.99 & 31.1 & 11298.1 & 412.28 & 213.7 & 29.1 & 63.4 & 63.6 & 53.3 & 8.66 & 67.00 & 3825.0 & 200.11 & 3724.27 & 33.9 & 63.4 & 64.7 & 54.4 \\
& & LoRA & 4.51 & 0.666 & 17.44 & 18.7 & 9399.3 & 414.87 & 129.3 & 54.0 & 63.2 & 63.0 & 53.3 & 8.66 & 75.00 & 3845.0 & 193.60 & 4033.33 & 57.0 & 62.5 & 64.0 & 53.6 \\
& & LoRA+ & 4.51 & 0.692 & 17.44 & 20.0 & 9143.4 & 389.40 & 129.8 & 54.6 & \cellcolor{red!15}\textbf{63.9} & 63.7 & 54.0 & 8.66 & 75.00 & 3854.0 & 192.00 & 4000.00 & 57.7 & 63.2 & 64.6 & 54.2 \\
& & \cellcolor{red!15}\textbf{QLoRA} & 4.51 & \textbf{0.703} & 17.44 & 25.5 & 8784.4 & \textbf{369.88} & 157.2 & \textbf{54.9} & \cellcolor{red!15}\textbf{63.9} & \cellcolor{blue!15}\textbf{64.0} & \textbf{54.1} & 8.66 & 92.00 & \textbf{2580.0} & \textbf{169.21} & 4324.26 & \textbf{58.0} & \textbf{63.4} & \textbf{65.3} & \textbf{54.9} \\
& & BitFit & \textbf{0.09} & 0.101 & \textbf{17.33} & \textbf{17.0} & \textbf{8606.9} & 415.76 & \textbf{117.8} & 38.2 & 30.6 & 30.3 & 20.7 & 8.66 & \textbf{49.00} & 3834.0 & 230.21 & \textbf{3133.41} & 41.3 & 30.0 & 31.2 & 21.1 \\
\cmidrule(lr){2-23}

& \multirow{5}{*}{Qwen3-1.7B}
& \cellcolor{red!15}\textbf{Full-FT} & 1720.57 & \textbf{0.723} & 43.00 & 51.6 & 20397.1 & 398.49 & 342.7 & 25.7 & \cellcolor{red!15}\textbf{63.6} & \cellcolor{blue!15}\textbf{63.6} & \textbf{52.8} & 14.33 & \textbf{76.00} & 4986.0 & 207.80 & \textbf{4386.89} & 30.4 & \textbf{63.9} & \textbf{65.1} & \textbf{54.6} \\
& & LoRA & 6.42 & 0.579 & 28.83 & 31.6 & 22318.8 & 406.71 & 214.2 & 47.8 & 60.2 & 59.6 & 49.4 & 14.33 & \textbf{76.00} & 5494.0 & 216.00 & 4560.00 & 50.9 & 60.0 & 61.2 & 50.6 \\
& & LoRA+ & 6.42 & 0.699 & 28.83 & 33.5 & 22799.5 & 388.30 & 216.8 & \textbf{51.1} & 63.5 & 62.9 & 52.6 & 14.33 & 86.00 & 5526.0 & 201.81 & 4821.02 & \textbf{54.1} & 63.2 & 64.4 & 53.8 \\
& & QLoRA & 6.42 & 0.612 & 28.83 & 40.3 & \textbf{18678.6} & \textbf{371.17} & 249.3 & 48.8 & 61.0 & 60.8 & 50.4 & 14.33 & 114.00 & \textbf{3952.0} & \textbf{183.33} & 5805.45 & 51.8 & 60.7 & 62.5 & 51.7 \\
& & BitFit & \textbf{0.12} & 0.043 & \textbf{28.67} & \textbf{29.5} & 21240.9 & 408.20 & \textbf{200.7} & 20.0 & 15.1 & 14.5 & 4.3 & 14.33 & 237.00 & 4954.0 & 205.63 & 13537.31 & 23.0 & 13.6 & 16.1 & 4.4 \\
\cmidrule(lr){2-23}

& \multirow{4}{*}{Mamba-1.4B}
& Full-FT & 1383.31 & 0.591 & 33.98 & 54.0 & 14758.6 & 426.78 & 384.1 & 24.1 & 60.0 & 60.4 & 49.5 & 11.33 & \textbf{371.00} & 4582.0 & 157.65 & 16246.71 & 28.9 & 58.9 & 61.7 & 49.8 \\
& & LoRA & 11.13 & 0.488 & 22.92 & 40.3 & 12291.6 & 411.96 & 276.7 & 43.5 & 57.0 & 57.3 & 46.8 & 11.33 & 652.00 & 4700.0 & \textbf{116.50} & 21099.44 & 46.5 & 55.3 & 58.4 & 46.2 \\
& & \cellcolor{red!15}\textbf{LoRA+} & 11.13 & \textbf{0.635} & 22.92 & 42.7 & 12347.2 & \textbf{396.39} & 282.1 & 48.0 & \cellcolor{red!15}\textbf{61.5} & \cellcolor{blue!15}\textbf{61.9} & \textbf{51.3} & 11.33 & 484.00 & 4690.0 & 140.48 & 18886.76 & 51.1 & \textbf{60.0} & \textbf{62.9} & \textbf{50.9} \\
& & BitFit & \textbf{0.39} & 0.367 & \textbf{22.65} & \textbf{34.6} & \textbf{11158.9} & 434.57 & \textbf{250.6} & \textbf{53.1} & 52.1 & 52.5 & 42.0 & 11.33 & 373.00 & \textbf{4244.0} & 153.61 & \textbf{15915.70} & \textbf{56.1} & 50.7 & 53.5 & 41.6 \\
\cmidrule(lr){2-23}

& \multirow{4}{*}{Mamba2-1.3B}
& Full-FT & 1351.87 & \textbf{0.576} & 35.79 & 32.4 & 17391.0 & 402.59 & 217.4 & 23.6 & \cellcolor{red!15}\textbf{60.1} & 59.8 & 49.5 & 11.93 & 384.00 & \textbf{4170.0} & 140.00 & 14933.33 & 28.3 & \textbf{58.6} & \textbf{61.4} & \textbf{49.5} \\
& & LoRA & 8.11 & 0.326 & 24.06 & \textbf{25.5} & 11626.2 & 431.06 & 183.2 & 37.6 & 50.4 & 50.4 & 40.3 & 11.93 & 540.00 & 4232.0 & \textbf{137.00} & 20550.00 & 40.7 & 48.4 & 51.5 & 39.3 \\
& & \cellcolor{red!15}\textbf{LoRA+} & 8.11 & 0.572 & 24.06 & 31.7 & 11514.6 & \textbf{365.30} & 193.0 & 47.4 & \cellcolor{red!15}\textbf{60.1} & \cellcolor{blue!15}\textbf{60.1} & \textbf{50.0} & 11.93 & 570.00 & 4345.0 & 142.00 & 22483.33 & 50.4 & 58.0 & 61.2 & 48.9 \\
& & BitFit & \textbf{0.21} & 0.326 & \textbf{23.86} & 27.7 & \textbf{10482.9} & 373.21 & \textbf{172.3} & \textbf{53.5} & 50.5 & 50.5 & 40.4 & 11.93 & \textbf{346.00} & 4300.0 & 155.00 & \textbf{14897.22} & \textbf{56.5} & 48.7 & 51.4 & 39.6 \\

\midrule
\midrule

\multirow{18}{*}{\textbf{LaMP-3}}
& \multirow{5}{*}{TinyLlama-1.1B}
& Full-FT & 1104.55 & 0.676 & 25.99 & 160.5 & 11523.2 & 412.71 & 1104.0 & 26.4 & 59.0 & 60.9 & 48.8 & 8.66 & 16.37 & 3964.0 & 306.00 & 83.47 & 31.2 & 61.8 & 62.0 & 52.8 \\
& & \cellcolor{red!15}\textbf{LoRA} & 4.51 & \textbf{0.611} & 17.44 & 96.3 & 8662.1 & 411.78 & 660.9 & 52.8 & \cellcolor{red!15}\textbf{60.2} & \cellcolor{blue!15}\textbf{61.9} & \textbf{50.4} & 8.66 & 17.45 & 3982.0 & 293.00 & 85.21 & 55.8 & \textbf{62.4} & 62.7 & 53.4 \\
& & LoRA+ & 4.51 & 0.618 & 17.44 & 103.6 & 8529.1 & 384.27 & 663.5 & 52.7 & 60.1 & 61.8 & 50.3 & 8.66 & 17.45 & 3982.0 & 294.02 & 85.51 & 55.7 & 62.4 & 62.6 & 53.4 \\
& & QLoRA & 4.51 & 0.631 & 17.44 & 129.8 & 7876.8 & \textbf{371.46} & 803.6 & 52.5 & 59.8 & 61.8 & 50.1 & 8.66 & 19.90 & \textbf{2582.0} & \textbf{265.19} & 87.95 & 55.6 & 62.2 & \textbf{63.0} & \textbf{53.7} \\
& & BitFit & \textbf{0.09} & 1.081 & \textbf{17.33} & \textbf{87.4} & \textbf{7866.9} & 411.56 & \textbf{599.5} & \textbf{65.3} & 55.9 & 57.5 & 46.1 & 8.66 & \textbf{15.15} & 3966.0 & 324.61 & \textbf{81.96} & \textbf{68.4} & 58.0 & 58.3 & 49.0 \\
\cmidrule(lr){2-23}

& \multirow{5}{*}{Qwen3-1.7B}
& Full-FT & 1720.57 & 0.626 & 43.00 & 270.2 & 20442.7 & 400.15 & 1802.0 & 22.9 & 59.0 & 60.8 & 48.2 & 14.33 & 17.75 & 5120.0 & 300.54 & 88.91 & 27.6 & 62.2 & 62.3 & 53.0 \\
& & LoRA & 6.42 & 0.631 & 28.83 & 165.6 & 15248.5 & 403.88 & 1114.7 & 48.8 & 59.4 & 61.0 & 49.0 & 14.33 & \textbf{17.50} & 5494.0 & 299.59 & \textbf{87.38} & 51.9 & 62.2 & 62.2 & 52.9 \\
& & \cellcolor{red!15}\textbf{LoRA+} & 6.42 & \textbf{0.614} & 28.83 & 175.6 & 16325.3 & 386.55 & 1131.3 & 49.0 & \cellcolor{red!15}\textbf{59.6} & \cellcolor{blue!15}\textbf{61.2} & \textbf{49.1} & 14.33 & 18.67 & 5526.0 & 285.73 & 88.89 & 52.0 & \textbf{62.4} & 62.3 & 53.0 \\
& & QLoRA & 6.42 & 0.633 & 28.83 & 215.7 & 14345.4 & \textbf{366.59} & 1317.9 & 48.8 & 59.2 & 61.1 & 48.8 & 14.33 & 22.53 & \textbf{3948.0} & \textbf{249.77} & 93.80 & 51.8 & 62.1 & \textbf{62.5} & \textbf{53.1} \\
& & BitFit & \textbf{0.12} & 3.093 & \textbf{28.67} & \textbf{153.7} & \textbf{14192.5} & 407.51 & \textbf{1043.9} & \textbf{50.2} & 43.5 & 45.1 & 33.1 & 14.33 & 25.72 & 5054.0 & 269.00 & 115.30 & \textbf{53.2} & 45.9 & 46.3 & 36.7 \\
\cmidrule(lr){2-23}

& \multirow{4}{*}{Mamba-1.4B}
& Full-FT & 1383.31 & 0.620 & 33.98 & 276.8 & 14751.9 & 426.74 & 1968.7 & 24.9 & 58.9 & 61.2 & 48.5 & 11.33 & 34.10 & 4450.0 & 219.96 & 125.01 & 29.7 & 61.9 & 62.5 & 52.8 \\
& & LoRA & 11.13 & 0.644 & 22.92 & 203.9 & 10747.7 & 410.44 & 1394.8 & 47.3 & 59.1 & 61.3 & 49.0 & 11.33 & 40.70 & 4366.0 & 201.52 & 136.70 & 50.4 & 61.6 & 62.3 & 52.5 \\
& & \cellcolor{red!15}\textbf{LoRA+} & 11.13 & \textbf{0.598} & 22.92 & 214.7 & 10998.6 & \textbf{397.39} & 1422.0 & 47.8 & \cellcolor{red!15}\textbf{59.5} & \cellcolor{blue!15}\textbf{61.8} & \textbf{49.4} & 11.33 & 40.85 & 4526.0 & \textbf{198.21} & 134.95 & 50.8 & \textbf{62.1} & \textbf{62.7} & \textbf{52.9} \\
& & BitFit & \textbf{0.39} & 0.665 & \textbf{22.65} & \textbf{173.2} & \textbf{9622.2} & 434.83 & \textbf{1255.2} & \textbf{61.7} & 59.0 & 61.2 & 49.0 & 11.33 & \textbf{33.85} & \textbf{4245.0} & 220.28 & \textbf{124.27} & \textbf{64.7} & 61.5 & 62.1 & 52.4 \\
\cmidrule(lr){2-23}

& \multirow{4}{*}{Mamba2-1.3B}
& Full-FT & 1351.87 & 0.581 & 35.79 & 169.2 & 17144.5 & 406.70 & 1146.9 & 25.2 & 59.9 & 61.5 & 49.4 & 11.93 & \textbf{32.75} & 4228.0 & 227.00 & 123.91 & 30.0 & 62.4 & 63.0 & 53.3 \\
& & \cellcolor{red!15}\textbf{LoRA} & 8.11 & \textbf{0.575} & 24.06 & \textbf{134.4} & 11063.5 & 429.02 & 961.0 & 49.2 & \cellcolor{red!15}\textbf{60.2} & \cellcolor{blue!15}\textbf{62.0} & \textbf{50.1} & 11.93 & 39.62 & 4169.0 & 196.00 & 129.41 & 52.3 & \textbf{62.4} & \textbf{63.1} & \textbf{53.3} \\
& & LoRA+ & 8.11 & 0.599 & 24.06 & 165.3 & 11150.6 & \textbf{368.86} & 1016.2 & 48.9 & 59.9 & 61.7 & 49.8 & 11.93 & 41.48 & \textbf{4068.0} & \textbf{188.00} & 129.98 & 52.0 & 62.1 & 62.8 & 53.1 \\
& & BitFit & \textbf{0.21} & 0.721 & \textbf{23.86} & 147.0 & \textbf{9902.2} & 376.20 & \textbf{921.7} & \textbf{63.6} & 58.7 & 60.6 & 48.7 & 11.93 & 35.03 & 4445.0 & 205.00 & \textbf{119.70} & \textbf{66.6} & 60.9 & 61.4 & 51.8 \\

\bottomrule
\end{tabular}%
}
\end{table*}

\section{Results and Analysis}
{\color{black}  



\subsection{Step 1: Best PEFT Method across Models}


For each model and benchmark, we select the finetuning method with a strict energy-first rule: the method with the highest finetuning NS-E is selected, and if two or more methods share the same NS-E, the one with the higher NS\# wins. NS-E measures energy efficiency using training time and power consumption, which is the binding constraint for battery-powered on-device fine-tuning. The memory-focused NS-M (peak VRAM) and the combined NS\# are still reported and discussed, since NS-M identifies the method to prefer when memory rather than energy is the dominant constraint, but only NS-E and the NS\# tie-break determine the selection. Although the parameter-based NS is reported in the tables for completeness, it is not used for method selection or for the following analysis because its strong dependence on the number of trainable parameters can favor methods with poor task performance. Table~\ref{tab:sst2_merged} reports the GLUE results, while Table~\ref{tab:lamp1_merged} reports the LaMP results.


\subsubsection{Perspective on PEFT Methods}
A key observation is that \textit{LoRA+ is by far the most frequently selected PEFT method}, being chosen in 18 of the 24 model--task pairs. Once energy alone decides the selection, the choice is also much less architecture-dependent than it is under a memory-aware rule: LoRA+ wins for both SSMs and for both Transformers on almost every task, and QLoRA, whose advantage lies in VRAM rather than energy, survives in a single configuration.
 
\textbf{LoRA+}, specifically, is ranked the top fine-tuning method in 3 out of 6 tasks for TinyLlama-1.1B,  
4 out of 6 for Qwen3-1.7B, 6 out of 6 for Mamba-1.4B, 
and 5 out of 6 for Mamba2-1.3B. 
One of the reasons why LoRA+ is the best method is that LoRA+ modifies only the optimizer and not the adapter itself. It keeps the same rank, the same trainable-parameter count and the same FLOPs as LoRA, and simply assigns a larger learning rate to the up-projection $B$ than to the down-projection $A$~\cite{hayou2024loraefficientlowrank}. Any accuracy it gains is therefore obtained at essentially zero additional cost, and since NetScore multiplies a performance term by cost terms that remain unchanged, the gain translates directly into a higher NS-E and NS-M. This is visible in the tables, where LoRA and LoRA+ share identical parameter counts and TFLOPs and differ only marginally in time, VRAM and power (e.g., Mamba2-1.3B on SST-2: 34.24 vs.\ 31.70~min and 12{,}903 vs.\ 12{,}805~MB), while accuracy increases from 0.955 to 0.967. Secondly, because $B$ is initialized to zero while $A$ is random, a single shared learning rate under-trains $B$ relative to $A$; the enlarged learning rate on $B$  mainly accelerates convergence. Under the short training budget used here, faster convergence appears as higher accuracy at the same time and energy, which is exactly what NS-E rewards. Third, the benefit is largest where the pretrained model is furthest from the target task: the largest LoRA$\rightarrow$LoRA+ jumps occur on the personalization benchmarks and on the SSMs (Mamba2-1.3B on LaMP-2, 0.326 $\rightarrow$ 0.572), whereas TinyLlama on SST-2 is already near ceiling and gains nothing (0.959 vs.\ 0.957). 

\textbf{LoRA}, the standard PEFT method, is selected in 4 of the 24 pairs, always with an almost negligible NS-E difference from LoRA+: SST-2 for TinyLlama, LaMP-1 for Qwen3, and LaMP-3 for both TinyLlama and Mamba2. For example, on SST-2 TinyLlama reaches an NS-E of 70.0 with LoRA against 69.9 with LoRA+, and on LaMP-3 it reaches 60.2 against 60.1. 
%

\textbf{Full-FT} is selected in only 1 of the 24 model--task pairs. On LaMP-2, Qwen3-1.7B attains its highest accuracy (0.723) with Full-FT, which is enough to give it the highest NS-E of that group (63.6, against 63.5 for LoRA+) even though its base NS collapses to 25.7 under the parameter penalty. A second near-miss is TinyLlama on STS-B, where the whole finetuning run lasts under five minutes (4.75~min) so the energy penalty of updating all parameters nearly disappears; Full-FT gives the best correlation (0.909) and the highest NS-M of that group (68.1), but LoRA+ still takes the higher NS-E (70.6 vs.\ 70.2) and is therefore selected under the energy-first rule. Outside these cases full finetuning is never cost-efficient: from Table~\ref{tab:lamp1_merged}, Qwen3-1.7B on LaMP-1 achieved a lower accuracy of 0.739 under full fine-tuning versus 0.760 under LoRA, while consuming 366.1~Wh against 210.0~Wh. This trend is seen in 23 of the 24 model--task pairs, showing that updating all parameters is generally not cost-efficient at this model size.

\textbf{BitFit}, which updates only the bias terms in the fine-tuned model, failed to generate competitive results for any of the models examined and is never selected. The largest declines were found on STS-B (Mamba2: 0.023 correlation), LaMP-2 (Qwen3: 0.043 accuracy), and LaMP-3 (Qwen3: 3.093 RMSE), demonstrating that BitFit is not the optimal choice for fine-tuning model in general understanding and user personalization tasks. On STS-B,  BitFit even produces a negative NS\# ($-3.6$) for Mamba2-1.3B, confirming that bias-only adaptation cannot fit the regression objective.

\textbf{QLoRA}, available only for the Transformer models, trades training time for memory. On SST-2 it reduces TinyLlama's peak finetuning VRAM from 16{,}480~MB (LoRA) to 4{,}225~MB (Table~\ref{tab:sst2_merged}), but the de-quantization overhead lengthens training from 19.2 to 31.1~min and raises energy from 83.6 to 147.3~Wh. It therefore almost never achieves the best NS-E within a model group (its only case is a tie with LoRA+ at 63.9 for TinyLlama on LaMP-2), yet it consistently achieves the best NS-M: it holds the highest NS-M for both Transformers on SST-2 (70.1 and 69.5) and QNLI (69.3 and 68.8), as well as for TinyLlama on LaMP-2 (64.0), that is, in 5 of the 12 Transformer model--task pairs. Under the energy-first rule this memory advantage no longer drives the selection, so QLoRA is the selected method in only 1 of those 12 pairs, namely TinyLlama on LaMP-2, where it ties LoRA+ on NS-E and wins the NS\# tie-break (54.1 vs.\ 54.0). QLoRA thus remains the method of choice only when VRAM, not energy, is the binding constraint. The reason why QLoRA is unavailable for Mamba models is because SSM blocks  have highly sensitive feature maps within the selective scan mechanism and massive outliers in output activations not present in Transformer attention modules~\cite{chiang2024quambaposttrainingquantizationrecipe}, causing standard quantization to fail.
\subsubsection{Perspective on Models}
A key observation is that \textit{SSM-based models show a greater improvement from LoRA to LoRA+ than the Transformer-based models}. 

\textbf{Transformer Models}: 
For Transformer-based models, the gap between LoRA and LoRA+ is much smaller than for SSMs. On TinyLlama-1.1B, standard LoRA already matches or slightly exceeds LoRA+ on several tasks (e.g., an NS-E of 70.0 vs.\ 69.9 on SST-2 in Table~\ref{tab:sst2_merged}, and 60.2 vs.\ 60.1 on LaMP-3 in Table~\ref{tab:lamp1_merged}), while Qwen3-1.7B shows only marginal LoRA+ gains on SST-2 (69.5 vs.\ 69.6) and none on QNLI (67.5 for both). Transformers also uniquely support QLoRA, which is decisive whenever VRAM is weighted: it holds the highest NS-M for both Transformers on SST-2 and QNLI and for TinyLlama on LaMP-2, where it is also the best configuration on the base NetScore (NS 54.9, Table~\ref{tab:lamp_summary_merged}) and the only pair in which it is the selected method under the energy-first rule. Overall, Transformer models require less PEFT specialization than SSMs to reach competitive NetScore values under the same training budget. 

\textbf{SSM Models}: On LaMP-1, similar to LaMP-2 and LaMP-3, Mamba-1.4B improves from an accuracy of 0.570 with LoRA to 0.698 with LoRA+, a 22\% gain, while TinyLlama improves from an accuracy of 0.730 to an accuracy of 0.732. 
%
%
%
This aligns with recent findings that PEFT methods are generally more effective for Mamba than for Transformers~\cite{yoshimura2025mambapeftexploringparameterefficientfinetuning} and that Mamba exhibits greater stability under PEFT~\cite{halloran2025mambastatespacemodelslyapunovstable}. 

We speculate that two properties of Mamba explain this. First, the adapters are attached to the projections that feed the selective scan, which controls how information is carried across the whole sequence. Therefore, a small number of trainable weights can change the behaviour of the model much more than the same number of weights inside an attention block~\cite{yoshimura2025mambapeftexploringparameterefficientfinetuning}. Second, Mamba has been shown to stay stable when adapted with PEFT~\cite{halloran2025mambastatespacemodelslyapunovstable}, allowing it to absorb larger weight updates without destabilizing the training process. This is exactly what LoRA+ does with its larger learning rate, which would explain why moving from LoRA to LoRA+ helps the SSMs far more (up to $+75\%$ relative accuracy for Mamba2-1.3B on LaMP-2) than the Transformers, where the gain is a few percent at most. As SSM-specific PEFT methods mature~\cite{galim2025parameterefficientfinetuningstatespace}, the gap between SSM and Transformer-based models may decrease.

\subsubsection{Perspective on Benchmarks}
A key observation is that \textit{the benchmark family and not the fine-tuning method, sets the efficiency level that can be reached, while LoRA+ is a safer default on GLUE than on LaMP}. Every selected GLUE configuration attains a higher finetuning NS-E (66.7--70.6, Table~\ref{tab:glue_summary_merged}) than every selected LaMP configuration (59.5--65.0, Table~\ref{tab:lamp_summary_merged}), a separation of 7.9\% between the two maxima that no choice of method is able to close, because the LaMP tasks are generative, use longer inputs (512 versus 128 tokens) and therefore consume far more time and energy per run.

Within each family however, the winning method is not equally consistent. LoRA+ is the selected method in 11 of the 12 GLUE configurations (all four models on QNLI and STS-B, and both SSMs together with Qwen3-1.7B on SST-2), the single exception being TinyLlama-1.1B on SST-2, and by the smallest possible margin (NS-E 70.0 for LoRA against 69.9 for LoRA+). On LaMP it is selected in only 7 of the 12 configurations: plain LoRA takes three of the remaining slots (Qwen3-1.7B on LaMP-1, TinyLlama-1.1B and Mamba2-1.3B on LaMP-3), while QLoRA and Full-FT take one each on LaMP-2. The short, high-accuracy GLUE classification runs leave little room for accuracy differences between methods, so the small but consistent convergence benefit of LoRA+ decides the outcome, whereas the much longer generative LaMP runs produce larger accuracy spreads in which plain LoRA, QLoRA, or even Full-FT can occasionally take the highest NS-E.



We present the full per-task breakdown across three benchmarks (SST2, QNLI, STS-B) in
Tables~\ref{tab:sst2_merged} and the three personalization benchmarks
(LaMP-1, LaMP-2, LaMP-3) in Table~\ref{tab:lamp1_merged}.
Each table reports efficiency (Time, VRAM, Power, Energy), task performance, and the four
NetScore variants (NS, NS-E, NS-M, NS\#) for both finetuning and inference. For each model,
the optimal tuning method (the one with the highest NS-E, with NS\# breaking ties) is marked
in \textbf{bold} with a red-shaded method name. Below we analyze each detailed table in turn.

\textbf{GLUE Tasks}: 
\noindent\textbf{SST-2.}
On this sentiment-classification task, all four models reach high accuracy under LoRA and LoRA+ (0.955--0.967), with Mamba2-1.3B+LoRA+ attaining the single highest accuracy (0.967). Because SST-2 is comparatively easy, even bias-only BitFit stays competitive for the Transformers (TinyLlama 0.920, Qwen3 0.900) and therefore takes the top base NetScore (NS 77.0 and 73.7) owing to its negligible number of trainable parameters (0.09--0.12 million). However, BitFit performs poorly on Mamba2-1.3B (0.500 accuracy), showing that bias-only tuning is unreliable for SSMs even on relatively easy tasks. Once the objective shifts to energy (NS-E), LoRA and LoRA+ overtake BitFit for every model. QLoRA shows the expected memory benefit, cutting TinyLlama's finetuning VRAM from 16{,}480~MB (LoRA) to 4{,}225~MB, but its de-quantization overhead lengthens training time (19.2~min $\rightarrow$ 31.1~min) and raises energy, so it never leads on NS or NS-E even though it takes the highest NS-M for both Transformers (70.1 and 69.5); under the energy-first rule it is therefore not selected for either, the choice falling to LoRA for TinyLlama (NS-E 70.0) and to LoRA+ for Qwen3 (69.6). Inference VRAM is 1.3--5.9$\times$ smaller than finetuning across all configurations, the smallest ratio being QLoRA, which is already memory-frugal during training.
\noindent\textbf{QNLI.}
QNLI is the longest-running GLUE task, which further increases the energy gap between architectures. LoRA+ gives the best accuracy for three of the four models (0.946, 0.928, 0.926), the exception being TinyLlama, where Full-FT is marginally ahead (0.934 vs.\ 0.932), while BitFit again wins the base NetScore (e.g., TinyLlama NS 75.4) but at a steep accuracy cost (TinyLlama 0.839, Mamba2-1.3B 0.500). The training-time difference is noticeable: TinyLlama finetunes in $\approx$78--81~min versus $\approx$148~min for Mamba-1.4B, so despite comparable VRAM the SSMs record lower NS-E. QLoRA again minimizes VRAM (TinyLlama 4{,}797~MB, Qwen3 11{,}114~MB) at the price of the longest training times (TinyLlama 124.6~min), reproducing the memory--energy trade-off observed on SST-2; because those long runs depress its NS-E, LoRA+ is instead the selected method for all four models on this task (67.9, 67.5, 66.7, and 67.1).
\noindent\textbf{STS-B.}
STS-B is the one GLUE task where BitFit fails completely: as a correlation (regression) objective, bias-only tuning yields near-zero scores (TinyLlama 0.310, Qwen3 0.316, Mamba-1.4B 0.146, Mamba2-1.3B 0.023) and even a negative NS\# ($-3.6$) for Mamba2-1.3B. Consequently LoRA+ becomes the best method per model on the base NetScore and on NS-E, unlike the other GLUE tasks where BitFit topped the base NS, and it is the selected method for all four models; the two Transformers are the exception on NS-M, where Full-FT retains the highest value (68.1 for TinyLlama and 67.3 for Qwen3) because the extremely short runs (4.75 and 12.63~min) leave little room for an efficiency penalty. LoRA+ also recovers most of the accuracy that plain LoRA loses (e.g., Mamba2-1.3B 0.705 $\rightarrow$ 0.847), highlighting the benefit of its differential learning rates. Because all configurations train in under 13~min, STS-B is the most efficient GLUE task and shows the largest NetScore separation between strong and weak methods.

\begin{table*}[h!]
\centering
\caption{\textsc{GLUE Summary for Best Training Configurations (finetuning + inference).} The performance metric (\textbf{Perf.}) is accuracy for SST-2 and QNLI and Pearson--Spearman correlation for STS-B; it is shared across both sections. The \textbf{Best Method} is selected using the finetuning results, including Full-FT, under a strict energy-first rule: the method with the highest NS-E is selected, and a tie on NS-E is broken by the higher NS\#. NS = NetScore: NS-E uses time and power; NS-M uses VRAM; NS\# uses VRAM, time, and power. All variants use the $1/8$ coefficient. Finetuning Time/Energy are in min/Wh; Inference Time/Energy are in s/mWh. \textbf{Bold} marks the best value within each task, computed separately for the finetuning and inference sections.}
\vspace{0.2cm}
\resizebox{\textwidth}{!}{
\begin{tabular}{llll c ccccccccc ccccccccc}
\toprule
\multirow{3}{*}{\textbf{Task}} & \multirow{3}{*}{\textbf{Model}} & \multirow{3}{*}{\textbf{Architecture}} & \multirow{3}{*}{\textbf{Best Method}} & \multirow{3}{*}{\textbf{Perf.}} & \multicolumn{9}{c}{\textbf{Finetuning}} & \multicolumn{9}{c}{\textbf{Inference}} \\
\cmidrule(lr){6-14} \cmidrule(lr){15-23}
& & & & & \textbf{TFLOPs $\downarrow$} & \textbf{Time $\downarrow$} & \textbf{VRAM $\downarrow$} & \textbf{Power $\downarrow$} & \textbf{Energy $\downarrow$} & \textbf{NS $\uparrow$} & \textbf{NS-E $\uparrow$} & \textbf{NS-M $\uparrow$} & \textbf{NS\# $\uparrow$} & \textbf{TFLOPs $\downarrow$} & \textbf{Time $\downarrow$} & \textbf{VRAM $\downarrow$} & \textbf{Power $\downarrow$} & \textbf{Energy $\downarrow$} & \textbf{NS $\uparrow$} & \textbf{NS-E $\uparrow$} & \textbf{NS-M $\uparrow$} & \textbf{NS\# $\uparrow$} \\
& & & & & (TF) & (min) & (MB) & (W) & (Wh) & & & & & (TF) & (s) & (MB) & (W) & (mWh) & & & & \\
\midrule
\multirow{4}{*}{\textbf{SST-2}}
& TinyLlama-1.1B & \multirow{2}{*}{Transformer} & LoRA & 0.959 & \textbf{16.17} & \textbf{19.18} & 16479.8 & \textbf{261.43} & \textbf{83.57} & \textbf{60.6} & \textbf{70.0} & 68.7 & \textbf{59.5} & \textbf{8.03} & \textbf{18.73} & \textbf{3654.0} & \textbf{94.00} & \textbf{489.06} & \textbf{63.7} & \textbf{71.2} & \textbf{70.4} & \textbf{62.3} \\
& Qwen3-1.7B & & LoRA+ & 0.962 & 23.49 & 22.30 & 20047.6 & 350.26 & 130.18 & 57.5 & 69.6 & 68.6 & 58.8 & 11.67 & 23.00 & 4678.0 & 101.00 & 645.28 & 60.6 & 70.9 & 70.2 & 61.7 \\
\cmidrule(lr){2-23}
& Mamba-1.4B & \multirow{2}{*}{SSM} & LoRA+ & 0.958 & 21.24 & 49.37 & \textbf{10779.6} & 383.87 & 315.86 & 55.5 & 68.6 & \textbf{69.2} & 58.5 & 10.48 & 23.50 & 3970.0 & 105.00 & 685.42 & 58.6 & 70.8 & 70.3 & 61.8 \\
& Mamba2-1.3B & & LoRA+ & \textbf{0.967} & 22.38 & 31.70 & 12805.1 & 357.50 & 188.88 & 56.8 & 69.3 & 69.1 & 59.0 & 11.09 & 31.70 & 4356.0 & 104.00 & 915.78 & 59.9 & 70.6 & 70.3 & 61.5 \\
\midrule
\multirow{4}{*}{\textbf{QNLI}}
& TinyLlama-1.1B & \multirow{2}{*}{Transformer} & LoRA+ & 0.932 & \textbf{16.17} & \textbf{81.46} & \textbf{16728.6} & \textbf{283.19} & \textbf{384.49} & \textbf{60.1} & \textbf{67.9} & \textbf{68.2} & \textbf{57.3} & \textbf{8.03} & 153.00 & \textbf{3158.0} & \textbf{86.53} & \textbf{3677.53} & \textbf{63.2} & 68.5 & \textbf{70.0} & \textbf{59.7} \\
& Qwen3-1.7B & & LoRA+ & \textbf{0.946} & 23.49 & 104.20 & 23605.8 & 391.14 & 679.28 & 57.3 & 67.5 & 68.1 & 56.6 & 11.67 & \textbf{144.00} & 5668.0 & 99.50 & \textbf{3980.00} & 60.3 & \textbf{68.6} & 69.7 & 59.3 \\
\cmidrule(lr){2-23}
& Mamba-1.4B & \multirow{2}{*}{SSM} & LoRA+ & 0.928 & 21.24 & 147.90 & 19639.4 & 421.40 & 1038.76 & 55.0 & 66.7 & 68.0 & 56.0 & 10.48 & 148.00 & 3942.0 & 97.61 & 4012.86 & 58.0 & 68.3 & 69.7 & 59.3 \\
& Mamba2-1.3B & & LoRA+ & 0.926 & 22.38 & 101.35 & 17726.1 & 406.87 & 687.27 & 56.1 & 67.1 & 68.0 & 56.5 & 11.09 & 197.00 & 4390.0 & 98.75 & 5403.82 & 59.1 & 67.9 & 69.6 & 58.8 \\
\midrule
\multirow{4}{*}{\textbf{STS-B}}
& TinyLlama-1.1B & \multirow{2}{*}{Transformer} & LoRA+ & \textbf{0.893} & \textbf{16.17} & \textbf{2.56} & 19197.1 & 367.73 & \textbf{15.69} & \textbf{59.4} & \textbf{70.6} & \textbf{67.3} & \textbf{59.9} & \textbf{8.03} & \textbf{33.00} & 5676.0 & 102.38 & \textbf{938.48} & \textbf{62.4} & \textbf{69.2} & \textbf{68.6} & \textbf{59.8} \\
& Qwen3-1.7B & & LoRA+ & \textbf{0.893} & 23.49 & 3.42 & 23622.6 & \textbf{358.60} & 20.44 & 56.2 & 70.3 & 67.1 & 59.4 & 11.67 & 41.00 & 9536.0 & 118.15 & 1345.60 & 59.3 & 68.8 & 68.1 & 58.9 \\
\cmidrule(lr){2-23}
& Mamba-1.4B & \multirow{2}{*}{SSM} & LoRA+ & 0.869 & 21.24 & 5.18 & 18798.0 & 418.61 & 36.14 & 53.8 & 69.2 & 66.9 & 58.5 & 10.48 & 43.00 & 4526.0 & 104.00 & 1242.22 & 56.9 & 68.4 & 68.4 & 59.3 \\
& Mamba2-1.3B & & LoRA+ & 0.847 & 22.38 & 3.54 & \textbf{14986.6} & 381.36 & 22.50 & 54.5 & 69.3 & 66.7 & 58.9 & 11.09 & 52.00 & \textbf{4488.0} & \textbf{102.00} & 1473.33 & 57.6 & 67.8 & 68.0 & 58.7 \\
\bottomrule
\end{tabular}
}
\label{tab:glue_summary_merged}
\end{table*}

\begin{table*}[h!]
\centering
\caption{\textsc{LaMP Summary for Best Training Configurations (finetuning + inference).} The performance metric (\textbf{Perf.}) is accuracy for LaMP-1 and LaMP-2 and RMSE ($\downarrow$) for LaMP-3; it is shared across both sections. The \textbf{Best Method} is selected using the finetuning results, including Full-FT, under a strict energy-first rule: the method with the highest NS-E is selected, and a tie on NS-E is broken by the higher NS\#. NS = NetScore: NS-E uses time and power; NS-M uses VRAM; NS\# uses VRAM, time, and power. All variants use the $1/8$ coefficient. Finetuning Time/Energy are in min/Wh. Inference Energy is in mWh, while inference Time units are given inline (min for LaMP-1/LaMP-3 and s for LaMP-2). \textbf{Bold} marks the best value within each task, computed separately for finetuning and inference.}
\vspace{0.2cm}
\resizebox{\textwidth}{!}{
\begin{tabular}{llll c ccccccccc ccccccccc}
\toprule
\multirow{3}{*}{\textbf{Task}} & \multirow{3}{*}{\textbf{Model}} & \multirow{3}{*}{\textbf{Architecture}} & \multirow{3}{*}{\textbf{Best Method}} & \multirow{3}{*}{\textbf{Perf.}} & \multicolumn{9}{c}{\textbf{Finetuning}} & \multicolumn{9}{c}{\textbf{Inference}} \\
\cmidrule(lr){6-14} \cmidrule(lr){15-23}
& & & & & \textbf{TFLOPs $\downarrow$} & \textbf{Time $\downarrow$} & \textbf{VRAM $\downarrow$} & \textbf{Power $\downarrow$} & \textbf{Energy $\downarrow$} & \textbf{NS $\uparrow$} & \textbf{NS-E $\uparrow$} & \textbf{NS-M $\uparrow$} & \textbf{NS\# $\uparrow$} & \textbf{TFLOPs $\downarrow$} & \textbf{Time $\downarrow$} & \textbf{VRAM $\downarrow$} & \textbf{Power $\downarrow$} & \textbf{Energy $\downarrow$} & \textbf{NS $\uparrow$} & \textbf{NS-E $\uparrow$} & \textbf{NS-M $\uparrow$} & \textbf{NS\# $\uparrow$} \\
& & & & & (TF) & (min) & (MB) & (W) & (Wh) & & & & & (TF) & & (MB) & (W) & (mWh) & & & & \\
\midrule
\multirow{4}{*}{\textbf{LaMP-1}}
& TinyLlama-1.1B & \multirow{2}{*}{Transformer} & LoRA+ & 0.732 & \textbf{17.44} & \textbf{27.5} & \textbf{7608.5} & 324.22 & \textbf{148.6} & \textbf{55.6} & 64.7 & \textbf{64.9} & \textbf{55.0} & \textbf{8.66} & \textbf{12.52} min & \textbf{3966} & 313.50 & \textbf{65400.00} & \textbf{58.7} & 65.6 & 65.6 & 56.6 \\
& Qwen3-1.7B & & LoRA & \textbf{0.760} & 28.83 & 34.3 & 19097.9 & 367.35 & 210.0 & 52.6 & \textbf{65.0} & 64.5 & 54.3 & 14.33 & 13.33 min & 5458 & 305.00 & 67780.00 & 55.6 & \textbf{66.2} & \textbf{65.9} & \textbf{56.9} \\
\cmidrule(lr){2-23}
& Mamba-1.4B & \multirow{2}{*}{SSM} & LoRA+ & 0.698 & 22.92 & 50.8 & 11124.9 & 368.74 & 312.2 & 49.7 & 63.1 & 63.6 & 53.0 & 11.33 & 36.93 min & 4326 & 195.00 & 120030.00 & 52.7 & 64.1 & 64.7 & 55.0 \\
& Mamba2-1.3B & & LoRA+ & 0.722 & 24.06 & 78.0 & 7680.3 & \textbf{198.92} & 258.6 & 51.4 & 63.9 & 64.6 & 54.2 & 11.93 & 38.33 min & 4145 & \textbf{190.00} & 121390.00 & 54.5 & 64.7 & 65.3 & 55.6 \\
\midrule
\multirow{4}{*}{\textbf{LaMP-2}}
& TinyLlama-1.1B & \multirow{2}{*}{Transformer} & QLoRA & 0.703 & \textbf{17.44} & \textbf{25.5} & \textbf{8784.4} & 369.88 & \textbf{157.2} & \textbf{54.9} & \textbf{63.9} & \textbf{64.0} & \textbf{54.1} & \textbf{8.66} & 92.00 s & \textbf{2580} & 169.21 & \textbf{4324.26} & \textbf{58.0} & 63.4 & \textbf{65.3} & \textbf{54.9} \\
& Qwen3-1.7B & & Full-FT & \textbf{0.723} & 43.00 & 51.6 & 20397.1 & 398.49 & 342.7 & 25.7 & 63.6 & 63.6 & 52.8 & 14.33 & \textbf{76.00} s & 4986 & 207.80 & 4386.89 & 30.4 & \textbf{63.9} & 65.1 & 54.6 \\
\cmidrule(lr){2-23}
& Mamba-1.4B & \multirow{2}{*}{SSM} & LoRA+ & 0.635 & 22.92 & 42.7 & 12347.2 & 396.39 & 282.1 & 48.0 & 61.5 & 61.9 & 51.3 & 11.33 & 484.00 s & 4690 & \textbf{140.48} & 18886.76 & 51.1 & 60.0 & 62.9 & 50.9 \\
& Mamba2-1.3B & & LoRA+ & 0.572 & 24.06 & 31.7 & 11514.6 & \textbf{365.30} & 193.0 & 47.4 & 60.1 & 60.1 & 50.0 & 11.93 & 570.00 s & 4345 & 142.00 & 22483.33 & 50.4 & 58.0 & 61.2 & 48.9 \\
\midrule
\multirow{4}{*}{\textbf{LaMP-3}}
& TinyLlama-1.1B & \multirow{2}{*}{Transformer} & LoRA & 0.611 & \textbf{17.44} & \textbf{96.3} & \textbf{8662.1} & 411.78 & \textbf{660.9} & \textbf{52.8} & \textbf{60.2} & 61.9 & \textbf{50.4} & \textbf{8.66} & \textbf{17.45} min & \textbf{3982} & 293.00 & \textbf{85210.00} & \textbf{55.8} & \textbf{62.4} & 62.7 & \textbf{53.4} \\
& Qwen3-1.7B & & LoRA+ & 0.614 & 28.83 & 175.6 & 16325.3 & \textbf{386.55} & 1131.3 & 49.0 & 59.6 & 61.2 & 49.1 & 14.33 & 18.67 min & 5526 & 285.73 & 88890.00 & 52.0 & \textbf{62.4} & 62.3 & 53.0 \\
\cmidrule(lr){2-23}
& Mamba-1.4B & \multirow{2}{*}{SSM} & LoRA+ & 0.598 & 22.92 & 214.7 & 10998.6 & 397.39 & 1422.0 & 47.8 & 59.5 & 61.8 & 49.4 & 11.33 & 40.85 min & 4526 & 198.21 & 134950.00 & 50.8 & 62.1 & 62.7 & 52.9 \\
& Mamba2-1.3B & & LoRA & \textbf{0.575} & 24.06 & 134.4 & 11063.5 & 429.02 & 961.0 & 49.2 & \textbf{60.2} & \textbf{62.0} & 50.1 & 11.93 & 39.62 min & 4169 & \textbf{196.00} & 129410.00 & 52.3 & \textbf{62.4} & \textbf{63.1} & 53.3 \\
\bottomrule
\end{tabular}
}
\label{tab:lamp_summary_merged}
\end{table*}


\textbf{LaMP Tasks}: 
\textbf{LaMP-1.}
On personalized citation identification, plain LoRA achieves the highest accuracy for Qwen3-1.7B (0.760), while for TinyLlama LoRA+ is marginally ahead (0.732 vs.\ 0.730) and LoRA+ is clearly needed to reach higher results on the SSMs (Mamba-1.4B 0.570 $\rightarrow$ 0.698; Mamba2-1.3B 0.700 $\rightarrow$ 0.722). Unlike the GLUE tasks, BitFit remains moderately usable here (0.439--0.517 accuracy) and therefore still tops the base NetScore (e.g., TinyLlama NS 65.3), but LoRA/LoRA+ dominate once energy (NS-E) is considered. This is also the benchmark where Qwen3-1.7B overtakes TinyLlama at performance-weighted operating points, because its accuracy advantage (0.760 vs. 0.732) outweighs the energy penalty only when accuracy is heavily emphasized.
\noindent\textbf{LaMP-2.}
Personalized movie tagging exposes the largest LoRA$\rightarrow$LoRA+ gains: Qwen3-1.7B rises from 0.579 to 0.699, Mamba-1.4B from 0.488 to 0.635, and Mamba2-1.3B from 0.326 to 0.572. Here TinyLlama+QLoRA is the best configuration on the base NetScore (NS 54.9) while also reaching the top accuracy among the PEFT methods (0.703), a rare case where quantization beats other approaches. This is also the only task in the whole study where Full-FT is selected, for Qwen3-1.7B, whose 0.723 accuracy is enough to give it the highest NS-E of the group (63.6, against 63.5 for LoRA+) despite a base NS of only 25.7. BitFit is catastrophic for Qwen3-1.7B (0.043 accuracy, NS 20.0), confirming that bias-only tuning cannot handle the larger label space of this task. The generative nature of LaMP-2 also makes SSM inference expensive (Mamba-1.4B 484~s vs. TinyLlama 75~s), reducing the theoretical latency advantage of SSMs in practice.
\noindent\textbf{LaMP-3.}
Personalized product rating is measured by RMSE and is the most energy-intensive benchmark (599--1968~Wh of finetuning energy). LoRA and LoRA+ give the lowest RMSE for every model (e.g., TinyLlama 0.611, Mamba2-1.3B 0.575), whereas BitFit degrades poorly (Qwen3-1.7B RMSE 3.093), which pulls its NetScore down sharply once the accuracy term is applied. Because every configuration consumes a large amount of energy, the NetScore denominators are similar and the spread between models is the smallest among all six tasks. This makes LaMP-3 a low-separation benchmark in which secondary constraints (VRAM, latency) can reasonably drive the choice.
\subsection{Step 2: Best Model across PEFT Methods} 
\subsubsection{Model Perspective} 
Having selected the best method per model for each of the task in Step~1, Step~2 compares these four final configurations of the models with respect to their efficiency. The comparison connects the per-task summaries of the best training configurations in Table~\ref{tab:glue_summary_merged} (GLUE) and Table~\ref{tab:lamp_summary_merged} (LaMP) with the detailed per-configuration results in Tables~\ref{tab:sst2_merged} and~\ref{tab:lamp1_merged}.
\paragraph{Transformer-based models.}
TinyLlama-1.1B has the highest base NetScore (NS) among the best-method configurations on all six benchmarks (Tables~\ref{tab:glue_summary_merged} and~\ref{tab:lamp_summary_merged}): 60.6 on SST-2, 60.1 on QNLI, 59.4 on STS-B, 55.6 on LaMP-1, 54.9 on LaMP-2, and 52.8 on LaMP-3. STS-B deserves a note: the energy-first rule gives TinyLlama LoRA+ (NS-E 70.6) rather than Full-FT, which holds the highest NS-M of that group; had Full-FT been selected, the parameter penalty of updating all 1.1B weights would have cut TinyLlama's base NS to 34.1 and left Qwen3 (56.2) in front. TinyLlama leads or ties the energy-focused variant (NS-E) on five of the six benchmarks (70.0 on SST-2, 67.9 on QNLI, 70.6 on STS-B, 63.9 on LaMP-2, and 60.2 on LaMP-3, tied with Mamba2-1.3B), losing only on LaMP-1 to Qwen3 (65.0 vs.\ 64.7). It leads the memory-focused variant (NS-M) on four of the six benchmarks, losing on SST-2 to Mamba-1.4B with LoRA+ (69.2 vs.\ 68.7) and on LaMP-3 to Mamba2-1.3B with LoRA (62.0 vs.\ 61.9).
One of the reasons that explains why TinyLlama-1.1B outperformed other transformer-based model is that TinyLlama has fewer parameters (1.1 billion) than Qwen3 (1.7 billion), resulting in lower energy costs and lower VRAM requirements. 

In addition, TinyLlama has an optimized Transformer inference/training stack compared to SSM-based models. 
For instance, on STS-B (Table~\ref{tab:sst2_merged}), TinyLlama with LoRA+ achieved an NS of 59.4 and an NS-E of 70.6 at only 15.69 Wh, while Qwen3 with LoRA+ achieved an NS of 56.2 and an NS-E of 70.3 at 20.44 Wh—meaning that Qwen3 used 30\% more energy (with a 56\% larger model) to reach the same 0.893 correlation.
Qwen3-1.7B, despite its larger size and higher raw accuracy on several tasks, is penalized by the model-size and energy terms of the NetScore variants. Its only leading position is under the energy-focused NS-E on LaMP-1, where Qwen3 with LoRA reaches 65.0 versus 64.7 for TinyLlama with LoRA+ (Table~\ref{tab:lamp1_merged}), because its accuracy advantage of 0.760 versus TinyLlama's 0.732 is large enough to outweigh its 41\% higher training energy (210.0 vs. 148.6 Wh). 
This shows how the choice of NetScore variant can shift the optimal model: under the base NS, NS-M, and NS\# (size-, memory-, and all-inclusive priorities), TinyLlama wins even on LaMP-1, whereas under the energy-focused NS-E, Qwen3 prevails.
\paragraph{SSM-based models.}
Under their best LoRA-family configurations, Mamba2-1.3B outperformed Mamba-1.4B in NetScore values on five of the six tasks (all except LaMP-2) despite having fewer parameters, e.g., an NS-E of 69.3 versus 68.6 on SST-2 (Table~\ref{tab:sst2_merged}) and of 63.9 versus 63.1 on LaMP-1 (Table~\ref{tab:lamp1_merged}). 
On SST-2, Mamba2 achieved an accuracy of 0.967 -- the highest among all models -- while consuming 189 Wh, whereas Mamba-1.4B reached an accuracy of 0.958 with 316 Wh. Similarly, on LaMP-1, Mamba2 required only 259 Wh to reach a higher accuracy (0.722) than Mamba-1.4B, which consumed 312 Wh for an accuracy of 0.698. This shows a generation-to-generation improvement in SSM architectures.
However, neither Mamba model demonstrated the ability to keep pace with the Transformers in terms of base NetScore values on any of the six benchmarks: on STS-B, for instance, TinyLlama's selected LoRA+ configuration reaches a base NS of 59.4 against 54.5 for Mamba2-1.3B and 53.8 for Mamba-1.4B. The SSMs lead only on the memory-focused NS-M, and only twice: Mamba-1.4B on SST-2 (69.2) and Mamba2-1.3B on LaMP-3 (62.0). 
The main reason for this disparity lies in the training duration: on QNLI, Mamba-1.4B requires approximately 148 minutes to train, whereas TinyLlama requires just 81 minutes under the same LoRA+ configuration; similarly, on LaMP-3, Mamba-1.4B takes 215 minutes to train, whereas TinyLlama takes 104 minutes to train. 
These longer training durations can be attributed to a variety of factors related to the immature nature of the SSM software ecosystem, which contribute to increased energy utilization and reduced NetScore values. 


Although Mamba's selective scan has $O(N)$ complexity in the sequence length $N$, compared with $O(N^2)$ for self-attention~\cite{gu2024mambalineartimesequencemodeling}, Mamba-1.4B takes longer to train than the larger Qwen3-1.7B on five of the six tasks (49.4 versus 22.3 minutes on SST-2 and 147.9 versus 104.2 minutes on QNLI). It is also slower during inference on all six tasks (484 versus 76 seconds on LaMP-2), despite having about 18\% fewer parameters and a lower TFLOPs cost (21.2--22.9 TF versus 23.5--28.8 TF). This suggests that the difference comes from actual GPU use rather than theoretical complexity. One possible reason is that our sequence lengths (128 tokens on GLUE and 512 on LaMP) are too short for the quadratic cost of attention to become dominant. In addition, Mamba's sequential scan requires frequent memory access and cannot use GPU tensor cores as effectively as the dense matrix operations in Transformers~\cite{gu2024mambalineartimesequencemodeling}. Its beam-search decoding also uses less optimized kernels to copy and reorder the recurrent state at each step~\cite{chiang2024quambaposttrainingquantizationrecipe}. These are mainly software and implementation limits, as Mamba-2 uses a matrix-based scan that is reported to be 2--8$\times$ faster than Mamba~\cite{dao2024transformersssmsgeneralizedmodels} and trains in 31.7 versus 49.4 minutes on SST-2 in our results.

\subsubsection{VRAM Perspective}
While VRAM is not directly accounted for within the base NetScore, it is an important constraint for edge deployments. 
Mamba models most of the times offer a VRAM advantage over Transformer-based models of comparable capability: on LaMP-1, for example, Mamba2 utilizes only 7,680 MB of VRAM, essentially matching the much smaller TinyLlama (7,609 MB with LoRA+) while using only 40\% of the 19,098 MB required by Qwen3. 
Therefore, in environments where memory is constrained, SSM-based models may be preferred regardless of their lower NetScore values.
{\color{black}
Additionally, while number of parameters directly affects the VRAM used, it is also influenced by architectural differences, PEFT method used and specific task tested. For example, even though TinyLlama is the smallest model tested, intuitively it should use the least VRAM, however due to architecture differences, both Mamba models can use substaintially less VRAM even though they have more parameters. In STS-B benchmark, TinyLlama uses at most 24209 MB of VRAM, while Mamba1 and Mamba2 use at most 19158 and 16998 MB of VRAM respectively. Note, that this trend can change depending on the task, as for example on LaMP-3 benchmark task, TinyLlama manages to use on average the least VRAM compared to other models.
}
\subsubsection{Benchmark Perspective}
The relative rankings of models were consistent within each benchmark family. 
Across all three GLUE tasks, the base NetScore ranking of the best configurations is identical, TinyLlama $>$ Qwen3 $>$ Mamba2-1.3B $>$ Mamba-1.4B (60.6 $>$ 57.5 $>$ 56.8 $>$ 55.5 on SST-2, 60.1 $>$ 57.3 $>$ 56.1 $>$ 55.0 on QNLI, and 59.4 $>$ 56.2 $>$ 54.5 $>$ 53.8 on STS-B, Table~\ref{tab:glue_summary_merged}). STS-B joins this pattern only because the energy-first rule selects LoRA+ for TinyLlama; under a memory-weighted criterion Full-FT would be chosen there instead, dropping TinyLlama's base NS to 34.1 and placing it last. 
On the LaMP tasks, TinyLlama kept its lead on the base NetScore for all three benchmarks. Qwen3 remains second on LaMP-1 (52.6), but falls to last on LaMP-2 (25.7) because Full-FT is its selected method there, and Mamba2-1.3B moves ahead of Qwen3 into second place on LaMP-3 (52.8 $>$ 49.2 $>$ 49.0 $>$ 47.8, Table~\ref{tab:lamp_summary_merged}), reflecting the heavier size and energy penalties of Qwen3's 1.7B parameters on the personalization workloads.
%
LaMP-3 stands out as the task where model differences are smallest: among the best LoRA-family configurations, the finetuning NS-E values range only from 59.5 (Mamba-1.4B with LoRA+) to 60.2 (TinyLlama with LoRA and Mamba2-1.3B with LoRA), a spread of roughly 1\% (Table~\ref{tab:lamp1_merged}). 
Furthermore, LaMP-3 was found to be the most energy intensive task evaluated. It consumed between 661--1422 Wh across all four models, thus, due to the larger denominators used in the NetScore formula, the differences among scores are reduced and make selections regarding method/model less decisive for this specific benchmark.
{\color{black}
\subsubsection{Training Time Perspective}
Training time follows a consistent pattern across the six benchmarks: under a matched LoRA+ configuration, TinyLlama-1.1B is the fastest of the four models on every task. For example, on QNLI (Table~\ref{tab:sst2_merged}) TinyLlama requires only about 81 minutes to train, compared to 104 minutes for Qwen3-1.7B and 148 minutes for Mamba-1.4B; on LaMP-3 (Table~\ref{tab:lamp1_merged}) it takes 104 minutes against 176 minutes for Qwen3 and 215 minutes for Mamba-1.4B. This is mainly due to its lower parameter count and the more mature Transformer training stack compared to the SSM-specific kernels. Because the energy-first rule selects LoRA+ for all four models on QNLI, this ordering carries over directly to the selected configurations there (81.5 minutes for TinyLlama against 101.4 for Mamba2-1.3B and 147.9 for Mamba-1.4B), and it also survives on the tasks where the selected methods differ: on LaMP-2, TinyLlama with QLoRA still trains in 25.5 minutes against 31.7 for Mamba2-1.3B with LoRA+.
Within the SSM family, Mamba2-1.3B trains substantially faster than Mamba-1.4B on the GLUE tasks (31.7 vs. 49.4 minutes on SST-2, and 101 vs. 148 minutes on QNLI), which reflects the matrix-multiplication-based scan introduced in Mamba-2, replacing the sequential selective scan used in the original Mamba. However, this advantage is not preserved on the personalization tasks: on LaMP-1, Mamba2 (78.0 minutes) actually trains longer than Mamba-1.4B (50.8 minutes), showing that the gap between the two SSM generations also depends on the specific task and sequence length used.
When the training times are compared between the Transformer and SSM families, the gap is most visible on long-running tasks (QNLI, LaMP-3), and is very small on STS-B, where all four selected configurations finish training in under six minutes. Because the total energy consumption is essentially a product of training time and average power draw, these longer training durations of SSM-based models translate directly into reduced NetScore values, even when their VRAM footprint is competitive with the Transformer-based models.
\subsubsection{Power Perspective}
 The differences in average power draw between models during training are smaller than the differences in training time, however, the power draw is still a relevant factor in total energy consumption. On SST-2 and QNLI, TinyLlama-1.1B records the lowest average power draw: under a matched LoRA+ configuration it draws 269 W on SST-2 against 350 W for Qwen3-1.7B, 384 W for Mamba-1.4B and 358 W for Mamba2-1.3B; on QNLI it draws 283 W against 391 W, 421 W, and 407 W respectively. Mamba-1.4B has the highest average power draw on most of the GLUE tasks (384-421 W), which combined with its longer training time is the main reason for it having the highest energy consumption out of the four models.
The differences in power draw become noticeably smaller on LaMP-2 and LaMP-3, where nearly all configurations fall within the 365-435 W range. This is likely a result of the smaller per-device batch size (4 vs. 32) but longer sequence length (512 vs. 128) used for the LaMP experiments, which increases compute intensity per step and pushes the GPU to similar utilization levels regardless of architecture. LaMP-1 is the exception, spanning 187-417 W, mainly because Mamba2-1.3B runs at an unusually low draw on this task (199 W with LoRA+ and 187 W with BitFit) despite its long training time.
\subsubsection{Inference Analysis}
Across all model-task configurations, the amount of memory required for inference was significantly less than what was used for training. Inference memory ranged from 2,120 to 10,293 MB, a 23 to 84\% reduction compared to training memory, with the smallest reductions observed for QLoRA configurations whose training footprint is already quantized. For example, TinyLlama LoRA on SST-2 required only 3,654 MB for inference, while 16,480 MB was required for training.
There is a substantial difference in inference cost when comparing the GLUE classification task and the LaMP generative task. On the GLUE classification tasks, which produce one label or score for each input, all of the models took anywhere from seconds to just a couple minutes to finish an inference. Specifically, SST-2 took 15-34 seconds, STS-B took 21-58 seconds and QNLI took 112-213 seconds. The LaMP generative tasks, however, are much more costly as they  perform autoregressive generation with beam search (up to four beams and up to 128 tokens), while also having higher input size due to having in context the user profiles. On LaMP-1, inference took from about 12 minutes for TinyLlama up to 41 minutes for Mamba-1.4B. Similarly, inference on LaMP-3 took from 15 to 41 minutes. One of the most interesting observations is that even though SSM-based models have been theoretically shown to have linear time advantage, this does not hold true for autoregressive generation. For example, on LaMP-2, Mamba-1.4B performs an inference in 484 seconds, while TinyLlama performed the same inference in 75 seconds. 
The choice of fine-tuning method did not significantly affect the latency of inference for the Full-FT, LoRA and LoRA+ configurations, because the LoRA adapters are merged into the base weights prior to deployment and therefore introduce no additional overhead. However, QLoRA was found to always incur additional latency costs resulting from dequantizing adapters during inference, while having significant decrease in VRAM used during inference. As an example, on SST-2, TinyLlama QLoRA has a 76\% increase in inference latency compared to LoRA+ and a 14\% decrease in VRAM used.
}
\subsection{Step 3: Best PEFT--Model Combination across Benchmarks}
In this section, we find the best combination of model and method for each of the six benchmarks, connecting the per-model selections summarized in Table~\ref{tab:glue_summary_merged} (GLUE) and Table~\ref{tab:lamp_summary_merged} (LaMP) with the detailed per-configuration results in Tables~\ref{tab:sst2_merged} and~\ref{tab:lamp1_merged}.
The method inside the best combination depends on which efficiency dimension is prioritized. 
Under the base NetScore, which penalizes trainable parameters and FLOPs, BitFit forms the winning combination on four of the six benchmarks (SST-2, QNLI, LaMP-1, and LaMP-3) owing to its minimal 0.09--0.39\,M trainable parameters. The two exceptions are STS-B, where BitFit's near-zero correlation (e.g., 0.310 for TinyLlama) makes TinyLlama with LoRA+ the best pairing (NS 59.4, Table~\ref{tab:sst2_merged}), and LaMP-2, where TinyLlama with QLoRA combines the best PEFT accuracy (0.703) with reduced VRAM (NS 54.9, Table~\ref{tab:lamp1_merged}). 
Under the energy-focused NS-E, LoRA-based methods always form the best combination (e.g., TinyLlama with LoRA reaches 70.0 on SST-2 and TinyLlama with LoRA+ reaches 70.6 on STS-B), whereas under the memory-focused NS-M and the combined NS\#, QLoRA pairs best on SST-2, QNLI, and LaMP-2 (NS-M values of 70.1, 69.3, and 64.0, all with TinyLlama) owing to its 4-bit quantized base weights. 
Full-FT never forms the best combination under the base or energy-focused variants; its only competitive cases are on STS-B under NS-M (68.1) and NS\# (60.0, vs. 59.9 for LoRA+), where training takes under five minutes and the efficiency penalty nearly vanishes. Even there the energy-first rule prefers LoRA+ for TinyLlama, so the single configuration in which Full-FT is selected is Qwen3 on LaMP-2, where its high accuracy gives it the highest NS-E of the group. As such, we conclude that updating all parameters is generally not a cost-efficient choice at this model scale. 
From a practical standpoint, an organization deploying SLMs on edge devices can therefore select the fine-tuning method according to its dominant constraint: BitFit when the trainable-parameter budget matters most (and the task tolerates its accuracy drop), LoRA or LoRA+ when energy is the constraint, and QLoRA when VRAM is the constraint.
TinyLlama-1.1B dominates the final selection, providing the best combination on all six benchmarks under the base NetScore: TinyLlama with BitFit on SST-2 (77.0), QNLI (75.4), LaMP-1 (65.3), and LaMP-3 (65.3); TinyLlama with LoRA+ on STS-B (59.4); and TinyLlama with QLoRA on LaMP-2 (54.9). 
TinyLlama loses in two cases, but even then, it depends on the chosen priority. On LaMP-1, when energy is prioritized (NS-E), Qwen3 with LoRA wins with 65.0 versus 64.7 for TinyLlama with LoRA+ (a 0.5\% relative difference), since Qwen3's higher accuracy (0.760 vs. 0.732, a 3.7\% relative difference) outweighs its 41\% higher training energy (210.0 vs. 148.6 Wh); under the base NS, however, TinyLlama with LoRA+ still leads with 55.6 versus 52.6, a 5.4\% relative improvement. On LaMP-3, when memory is prioritized (NS-M), Mamba2 with LoRA (62.0) narrowly edges out TinyLlama with LoRA (61.9), a 0.2\% relative difference.
The six benchmarks can be grouped into three types based on how decisive the model-method selection is: 
\begin{enumerate}

\item \textit{High-Separation Benchmarks} (STS-B, LaMP-2): There is a very large difference in the NetScore values between the best and worst model-method pairs, driven mainly by BitFit collapses on these tasks: on STS-B, NS-E spans from 70.6 (TinyLlama with LoRA+) down to 6.7 (Mamba2 with BitFit), with NS\# even turning negative ($-3.6$); on LaMP-2, Qwen3 with BitFit falls to an NS-E of 15.1. High-separation benchmarks have high sensitivity to both model quality and efficiency, so choosing between configurations is very important.

\item \textit{Medium-Separation Benchmarks} (SST-2, QNLI, LaMP-1): There is some difference between the top and bottom model-method pairs in terms of their NetScore values (for example, on SST-2 the base NS of the best configurations ranges from 77.0 for TinyLlama down to 61.3 for Mamba2, a 20\% spread computed as $100\times(\text{max}-\text{min})/\text{max}$). In these cases, TinyLlama is a clear winner, however a user who has strict memory limitations may choose an SSM-based model at a small penalty in terms of NetScore. 
\item \textit{Low-Separation Benchmarks} (LaMP-3): The NetScore values of all reasonable configurations are close together (the finetuning NS-E of the best LoRA-family configurations ranges only from 59.5 to 60.2, a spread of about 1\%) since all configurations require significant amounts of energy (661--1422 Wh). For low-separation benchmarks, the choice of model and method has relatively little impact, thus users can prioritize other constraints such as VRAM and latency, without a significant loss in NetScore.
\end{enumerate}

}






{\color{black}
\section{Conclusion}

This paper investigated whether SLMs can be fine-tuned with Parameter-Efficient Fine-Tuning as a practical, energy-aware alternative to full fine-tuning of large language models. By assessing four models from two families (Transformer-based: TinyLlama-1.1B and Qwen3-1.7B; SSM-based: Mamba-1.4B and Mamba2-1.3B) with five fine-tuning methods (Full Fine-Tuning, LoRA, LoRA+, QLoRA, and BitFit) across six benchmarks (three GLUE and three LaMP), we reach two main conclusions. First, PEFT methods match or surpass full fine-tuning at substantially lower resource cost: BitFit achieves the highest base NetScore in 18 of 24 model--task configurations due to its minimal trainable parameters, while LoRA+ leads the energy-focused NetScore-E in 19 of 24 configurations by balancing accuracy with low training energy and is therefore the method selected by our energy-first rule in 18 of 24 configurations; full fine-tuning never ranks first on the base NetScore and is selected only once. Second, TinyLlama-1.1B is the most efficient model overall, providing the best model--method combination under the base NetScore on all six benchmarks, largely because of its compact size and the maturity of the Transformer training stack. The optimal PEFT method depends on the priority: BitFit for parameter efficiency, LoRA+ for energy, and QLoRA for memory. Although the Mamba models gain substantially from LoRA+, they remain limited by longer training times due to less mature SSM libraries and toolkits. These findings indicate that a practical path to sustainable on-device NLP lies in compact SLMs combined with PEFT, rather than ever-larger models. Limitations include evaluation on a single GPU (RTX 4090), the unavailability of QLoRA for Mamba models, and the fact that the base NetScore does not directly penalize VRAM. Future work should address these constraints and extend the study to edge hardware, hybrid Transformer--SSM architectures, and additional personalization benchmarks.
}

\bibliographystyle{IEEEtran}
\bibliography{ref}

\appendices

\end{document}